%% file: main.tex
\ifdefined\XeTeXversion\else\ifdefined\directlua\else\pdfoutput=1\fi\fi
\documentclass[11pt]{article}
\usepackage[numbers]{clain}
\input{preamble}

\hypersetup{
  pdftitle={Pothen: Can Vision-Language Models Tell Where a Cultural Object Comes From?},
  pdfauthor={Yuechen Jiang, Mohsinul Kabir, Stavroula Golfomitsou, Konstantinos Arvanitis, Sophia Ananiadou}
}

\IfFileExists{T1Lbstr-LF.fd}{%
}{%
}
\DeclareRobustCommand{\Pothen}{%
  \texorpdfstring{%
    {\color{PothenForest}\fontencoding{T1}\fontfamily{\PothenFontFamily}%
     \fontseries{b}\fontshape{it}\selectfont Pothen}%
  }{Pothen}%
}

\newcommand{\PothenWebsite}{https://carolyn-jiang.github.io/Pothen/}
\title{\Pothen{}: Can Vision-Language Models Tell Where a Cultural Object Comes From?}
\date{22 November, 2026}
\reportlabel{Technical Report}
\author{%
Yuechen Jiang\textsuperscript{1}, Mohsinul Kabir\textsuperscript{1},
Stavroula Golfomitsou\textsuperscript{2}, Konstantinos Arvanitis\textsuperscript{1},
Sophia Ananiadou\textsuperscript{1,3}%
}
\affiliation[1]{University of Manchester}
\affiliation[2]{Getty Conservation Institute}
\affiliation[3]{ELLIS Manchester}
\correspondence{\mbox{}}
\metadata[Hugging Face]{\href{https://huggingface.co/datasets/Carolyn-Jiang/Metadata-Inference}{Metadata-Inference}}
\metadata[Website]{\href{\PothenWebsite}{\PothenWebsite}}

\begin{document}
\begin{abstract}
A cultural object is a trace of the people who made it, and knowing who made it, where, and when shapes how it is catalogued, valued, and returned. Vision-language models now describe such objects for museum collections, yet whether they can recover this knowledge from an image has not been tested. We introduce \Pothen{}, a benchmark of 750 objects from the Metropolitan Museum of Art and the Getty in which a model predicts an object's title, culture, period, origin, and creator from its image alone, scored against curator-written catalogue records by an LLM judge calibrated on human annotation. Across nine models, mean field scores are 36.16\% to 49.82\%, but all five fields match for only 2.13\% to 7.87\% of objects. Images raise culture, period, and origin scores by 12.6 to 32.2 points for two models tested without them, and 83.52\% of creator matches occur alongside an incorrect field. Models read visual evidence and hold facts about objects but do not bind them into one attribution, and in our error analysis their attributions follow the style an object resembles rather than the record of its making.
\end{abstract}
\maketitle

\section{Introduction}

A cultural object is a trace of the people who made it~\cite{gell1998art,geertz1973interpretation}, and knowing who made it, where, and when shapes how it is understood and treated. Museums catalogue and exhibit cultural objects by this knowledge~\cite{baca2016metadata,harpring2010controlled}, and it decides how objects are valued and whether they are returned to the communities they were taken from~\cite{gramlich2017provenance,schuhmacher2024nazi,christen2011opening}. This knowledge is hard to read from appearance, because styles travel through trade and imitation, so a vase made in America can look Dutch. Experts read it by combining what they see with what they know~\cite{chatterjee2014neuroaesthetics,leder2004model}.

\begin{figure}[!htbp]
  \centering
  \includegraphics[width=0.96\linewidth,height=0.44\textheight,keepaspectratio]{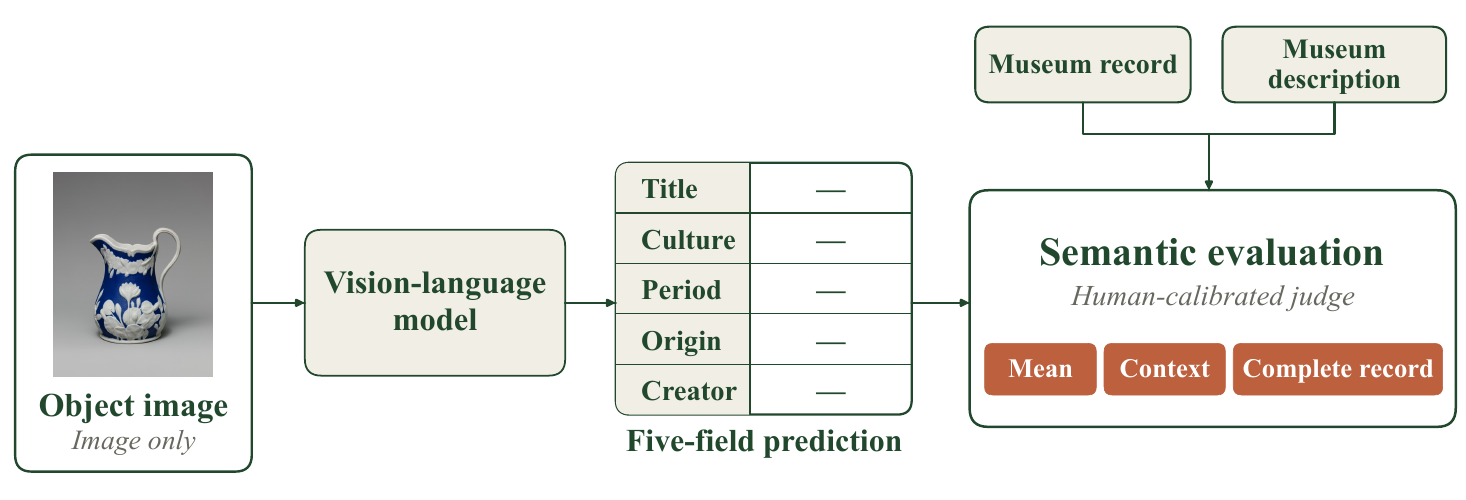}
  \caption{Image-only prediction of five metadata fields, evaluated against the museum record. Illustrative object image: \emph{Pitcher}, The Metropolitan Museum of Art (CC0). Dashes indicate schematic field values.}
  \label{fig:task-overview}
\end{figure}

\begin{table}[!t]
\caption{Heritage-related datasets. Who, When, Where: the benchmark evaluates prediction of an object's maker, date or period, and culture or origin. Open-ended: these attributes are predicted as free text. Record-level: a metric requires all attributes of an object to be jointly correct. Human Verif.: human-authored or human-verified annotations. Evaluated Models: output-producing models for pre-LLM resources. \cmark\ full, \pmark\ partial, \xmark\ absent.}
\label{tab:heritage_caption_comparison}
\centering
\fontsize{9}{11}\selectfont
\setlength{\tabcolsep}{2pt}
\begin{tabular*}{\textwidth}{@{\extracolsep{\fill}}>{\raggedright\arraybackslash}p{.16\textwidth}>{\centering\arraybackslash}m{.13\textwidth}cccccccc>{\centering\arraybackslash}m{.17\textwidth}@{}}
\toprule
\textbf{Dataset} & \textbf{Domain} & \textbf{\makecell{Multi-\\Culture}} & \textbf{\makecell{Multi-\\Type}} & \textbf{\makecell{Human\\Verif.}} & \textbf{Who} & \textbf{When} & \textbf{Where} & \textbf{\makecell{Open-\\ended}} & \textbf{\makecell{Record-\\level}} & \textbf{Evaluated Models} \\
\midrule
Artpedia~\cite{stefanini2019artpedia} & Painting & \pmark & \xmark & \cmark & \xmark & \xmark & \xmark & \xmark & \xmark & Retrieval (CNN + GRU) \\
MLF~\cite{liu2023feature} & Ceramics & \xmark & \xmark & \cmark & \xmark & \xmark & \xmark & \xmark & \xmark & Captioner (MLF) \\
CArt15K~\cite{zheng2023artalk} & Ceramics & \xmark & \xmark & \cmark & \xmark & \xmark & \xmark & \xmark & \xmark & Captioner (ARTalk) \\
GalleryGPT~\cite{bin2024gallerygpt} & Painting & \xmark & \xmark & \xmark & \xmark & \xmark & \xmark & \xmark & \xmark & MLLM \\
ArtCap~\cite{lu2024artcap} & Visual Art & \xmark & \xmark & \cmark & \xmark & \xmark & \xmark & \xmark & \xmark & Captioning models \\
DEArt~\cite{reshetnikov2025caption} & Painting & \xmark & \xmark & \cmark & \xmark & \xmark & \xmark & \xmark & \xmark & MLLM \\
ARTSEEK~\cite{fanelli2025artseek} & Visual Art & \xmark & \xmark & \xmark & \cmark & \xmark & \xmark & \xmark & \xmark & MLLM \\
MosAIC~\cite{bai2025power} & General Cultural & \cmark & \xmark & \pmark & \xmark & \xmark & \xmark & \xmark & \xmark & MLLM \\
DenseAnnotate~\cite{lin2025denseannotate} & General VLM & \cmark & \xmark & \cmark & \xmark & \xmark & \xmark & \xmark & \xmark & VLM \\
EmoArt~\cite{zhang2025emoart} & Painting & \cmark & \xmark & \pmark & \xmark & \xmark & \xmark & \xmark & \xmark & Text-to-image diffusion \\
Geo-TCAM~\cite{zhong2026geotcam} & Religious Painting & \xmark & \xmark & \cmark & \xmark & \xmark & \xmark & \xmark & \xmark & Captioner (ResNet-50 + Transformer) \\
Rijksmuseum Challenge~\cite{mensink2014rijksmuseum} & Museum Collection & \pmark & \cmark & \cmark & \cmark & \cmark & \xmark & \xmark & \xmark & Fisher vectors + SVM \\
OmniArt~\cite{strezoski2018omniart} & Visual Art & \pmark & \cmark & \cmark & \cmark & \cmark & \xmark & \xmark & \xmark & Multi-task CNN (ResNet-50) \\
iMet~\cite{zhang2019imet} & Museum Collection & \cmark & \cmark & \cmark & \xmark & \xmark & \pmark & \xmark & \xmark & Challenge classifiers \\
\midrule
\Pothen{} (ours) & \makecell{Multi-Category\\Heritage} & \cmark & \cmark & \cmark & \cmark & \cmark & \cmark & \cmark & \cmark & VLM \\
\bottomrule
\end{tabular*}

\end{table}

Vision-language models~\cite{radford2021clip,liu2023llava,bai2025qwen25vl} now tag and describe such objects for museum collections~\cite{villaespesa2021comparison,bin2024gallerygpt,reshetnikov2025caption}, often for readers who never see the objects themselves. For a human viewer, knowing an object's source changes how the same image is evaluated and how the brain responds to it~\cite{kirk2009modulation}. These models instead learn largely from paired images and text, which tie an object to what it resembles rather than to who made it. Whether they can tell who made an object, or only what it looks like, remains open.

We introduce \Pothen{}, the first benchmark that tests whether vision-language models can recover the full catalogue attribution of a cultural object from its image alone (Figure~\ref{fig:task-overview}). The dataset contains 750 objects from the Metropolitan Museum of Art and the Getty, drawn from four regions and four object types in fifteen strata, each paired with a catalogue record written by curators~\cite{baca2016metadata,harpring2010controlled}. The task asks a model, given only the image, to predict five fields of this record, namely title, culture, period, origin, and creator, as open-ended text rather than labels from a fixed vocabulary (Table~\ref{tab:heritage_caption_comparison}). For evaluation, an LLM judge calibrated against human annotation ($\kappa = 0.787$) scores each field, separately for fields the museum records and fields it leaves empty, and a record-level metric credits a prediction only when all fields match the same object, compared against an independent-field baseline. A control that gives only the museum name and title separates what models read from the image from what they recall.

Across nine models, mean field scores range from 36.16\% to 49.82\%, and all five fields match for only 2.13\% to 7.87\% of objects. These complete matches still exceed the independent-field baseline (0.57\% to 3.12\%), so models succeed on a few objects they know as a whole rather than reasoning through each field. Images raise culture, period, and origin scores by 12.6 to 32.2 points for the two models tested without them, yet 83.52\% of creator matches occur alongside an incorrect field. Models thus read visual evidence and hold facts about objects, but lack the binding between the two that the meaning-knowledge system supplies in human viewers~\cite{chatterjee2014neuroaesthetics,leder2004model}. One model names the Union Porcelain Works as the maker of a compote yet places it in European culture, following the object's style rather than the record of its making. \Pothen{} measures this binding directly through single fields, complete records, and image evidence.

\section{Related Work}
\textbf{Museum attribute prediction.}
The Rijksmuseum Challenge established artist, type, material, and creation-year prediction across museum objects~\cite{mensink2014rijksmuseum}. OmniArt predicts multiple artistic attributes with multi-task learning~\cite{strezoski2018omniart}, and iMet extends museum recognition to fine-grained culture and subject labels~\cite{zhang2019imet}. Recognition and retrieval methods also link images with catalogue metadata~\cite{saleh2016large,garcia2018read,garcia2019context,ypsilantis2021met}, and Strafforello et al. evaluate vision-language models on the style, author, and period of paintings~\cite{strafforello2024art}. These works classify paintings or objects into fixed label sets. \Pothen{} instead asks vision-language models for an open-ended attribution of five fields and credits it only when the fields agree on the same object, which tests whether a model's judgements cohere into one attribution.

\textbf{Heritage description and cultural evaluation.}
Artpedia distinguishes visual from contextual sentences in descriptions of paintings~\cite{stefanini2019artpedia}. Later work describes ceramics~\cite{liu2023feature,zheng2023artalk}, analyses artworks~\cite{lu2024artcap,reshetnikov2025caption,fanelli2025artseek,bin2024gallerygpt,lin2025denseannotate}, and models iconography and affect~\cite{cetinic2021iconographic,achlioptas2021artemis,zhang2025emoart} or culturally significant content~\cite{bai2025power,zhong2026geotcam}. Cultural benchmarks assess visual reasoning, commonsense, question answering, and critique~\cite{liu2021marvl,yin2021broaden,romero2024cvqa,nayak2024culturalvqa,maji2025drishtikon,liu2025culturevlm,blendvis2025,yu2026vulcabench,lee2025culturalbias}, and the notion of culture they measure varies widely~\cite{adilazuarda2024towards}. Free-text descriptions are hard to check against a record, and question answering tests what a model knows about a culture rather than whether it can place a specific object. \Pothen{} targets the catalogue record, so each attribution error can be traced to a field.

\textbf{Semantic evaluation of model outputs.}
Instruction-following VLMs support structured, image-conditioned prediction~\cite{li2023blip2,dai2023instructblip,chen2024internvl,llava_onevision,wang2024qwen2vl}. Caption metrics compare wording, semantic content, or image-text alignment~\cite{papineni2002bleu,vedantam2015cider,anderson2016spice,hessel2021clipscore}, whereas LLM judges can apply field-specific rubrics~\cite{zheng2023judging,liu2023geval,kim2024prometheus,chen2024mllmjudge}. Such judges carry biases and can differ from human assessment~\cite{wang2024fair,zheng2023judging}, and museum-tagging studies show that labels from vision models and descriptions by cataloguers often diverge~\cite{villaespesa2021comparison,villaespesa2021apple,villaespesa2021tags}. We therefore calibrate the judge's field decisions against human annotations and report ranking uncertainty.

\begin{figure}[t]
  \centering
  \includegraphics[width=\textwidth,height=0.70\textheight,keepaspectratio]{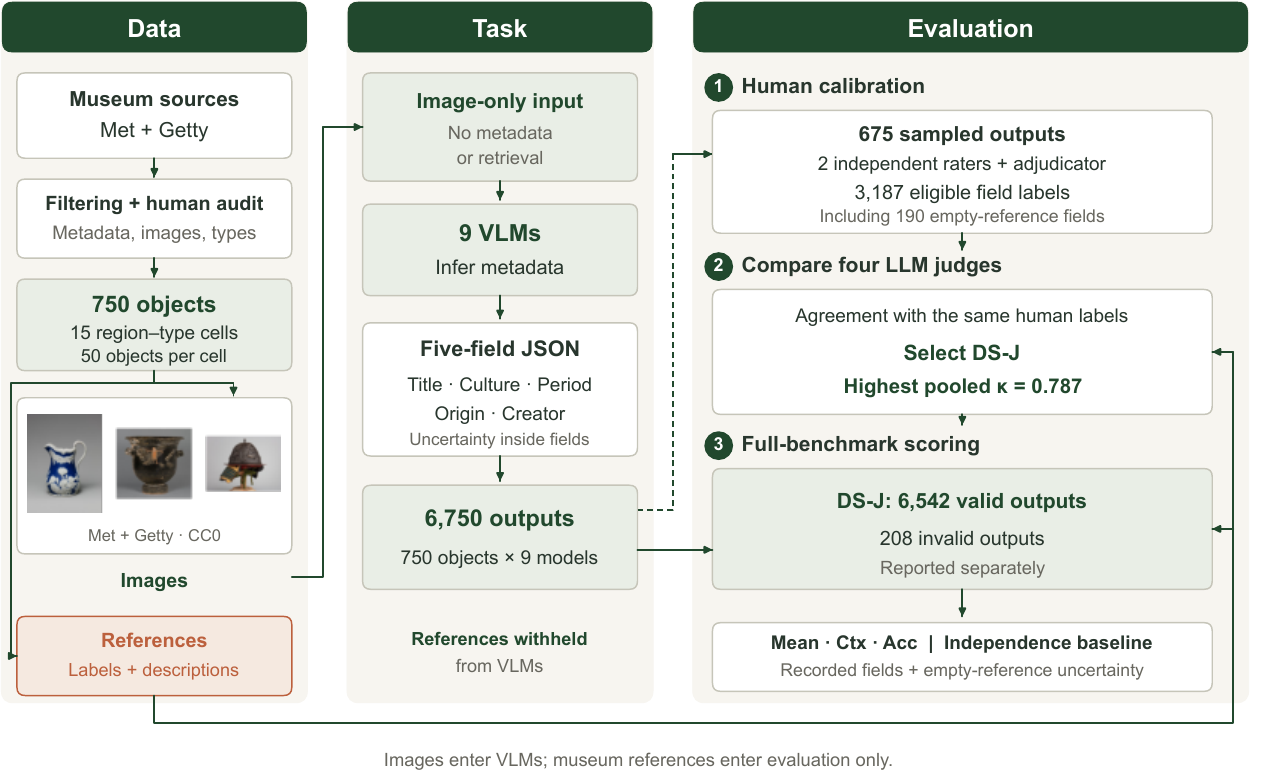}
  \caption{Benchmark workflow. Nine VLMs infer five-field metadata for 750 objects; the judge with the highest Cohen's $\kappa$ on 675 human-reviewed outputs scores all outputs. Judges compare predictions with reference labels and museum descriptions. Invalid outputs are reported separately. Illustrative collection images: Met and Getty (CC0).}
  \label{fig:workflow}
\end{figure}

\section{\Pothen{} Benchmark}

\Pothen{} tests whether a model can attribute a cultural object from its image alone. Figure~\ref{fig:workflow} summarises the workflow, and Appendix~\ref{appendix:metric-definitions} defines the metrics.

\subsection{Task Formulation}
\label{sec:task_formulation}

\paragraph{Problem Definition.}
For object $i$ with image input $X_i$, a VLM $f_\theta$ predicts $\hat{\mathcal{M}}_i=f_\theta(X_i)=\{\hat{m}_i^k\}_{k\in\mathcal{F}}$ over five fields $\mathcal{F}$ (title, culture, period, origin, and creator), evaluated against the museum record $\mathcal{M}_i^*=\{m_i^{*k}\}_{k\in\mathcal{F}}$. \emph{Title}, the recognition-oriented field, is the museum object name; \emph{culture} is the catalogued cultural affiliation; \emph{period} is a dated interval or named historical period; \emph{origin} is the recorded place of production or discovery; and \emph{creator} is the recorded person, workshop, manufacturer, or attributed group, including an explicitly unknown creator. Culture, period, origin, and creator together state who made an object, where, and when, and recovering them requires binding what is visible in the image to knowledge of the object's history. We call the task \emph{cultural attribution}; \emph{compatibility} denotes a positive field decision under the evaluation rubric. Text outside the five JSON fields is not scored. Field-level decisions are aggregated as $S\bigl(\hat{\mathcal{M}}_i,\mathcal{M}_i^*\bigr)=\frac{1}{|\mathcal{F}|}\sum_{k\in\mathcal{F}}c_{i,k}=s_i/|\mathcal{F}|$, where $c_{i,k}=J\bigl(\hat{m}_i^k,m_i^{*k}\bigr)\in\{0,1\}$ is the selected judge's field decision under the fixed rubric and museum description, $s_i$ counts compatible fields, and $m_i^{*k}$ contains the released reference-label alternatives. Nonempty references are scored for semantic compatibility; empty references follow Section~\ref{sec:evaluation}. Over valid outputs, mean field accuracy (Mean) averages $S$, context accuracy (Ctx) is the fraction with all four contextual decisions equal to 1, and complete-record accuracy (Acc) is the fraction with $S=1$. 

\subsection{Dataset Construction}
\label{sec:dataset-construction}

Objects were retrieved from the Metropolitan Museum of Art\footnote{\url{https://www.metmuseum.org/art/collection/}. Public-domain images and basic collection data are available under CC0.} and Getty\footnote{\url{https://www.getty.edu/art/collection/}. Public-domain Open Content images are under CC0; released IIIF metadata is under CC BY 4.0, subject to object-level rights statements.}, which publish curator-written catalogue records with open images. Candidates were retrieved with region keywords and four type filters (ceramics, paintings, metalwork, and sculpture), and regions define the sampling strata.

\begin{figure}[!htbp]
    \centering
    \includegraphics[width=0.92\linewidth,height=0.70\textheight,keepaspectratio]{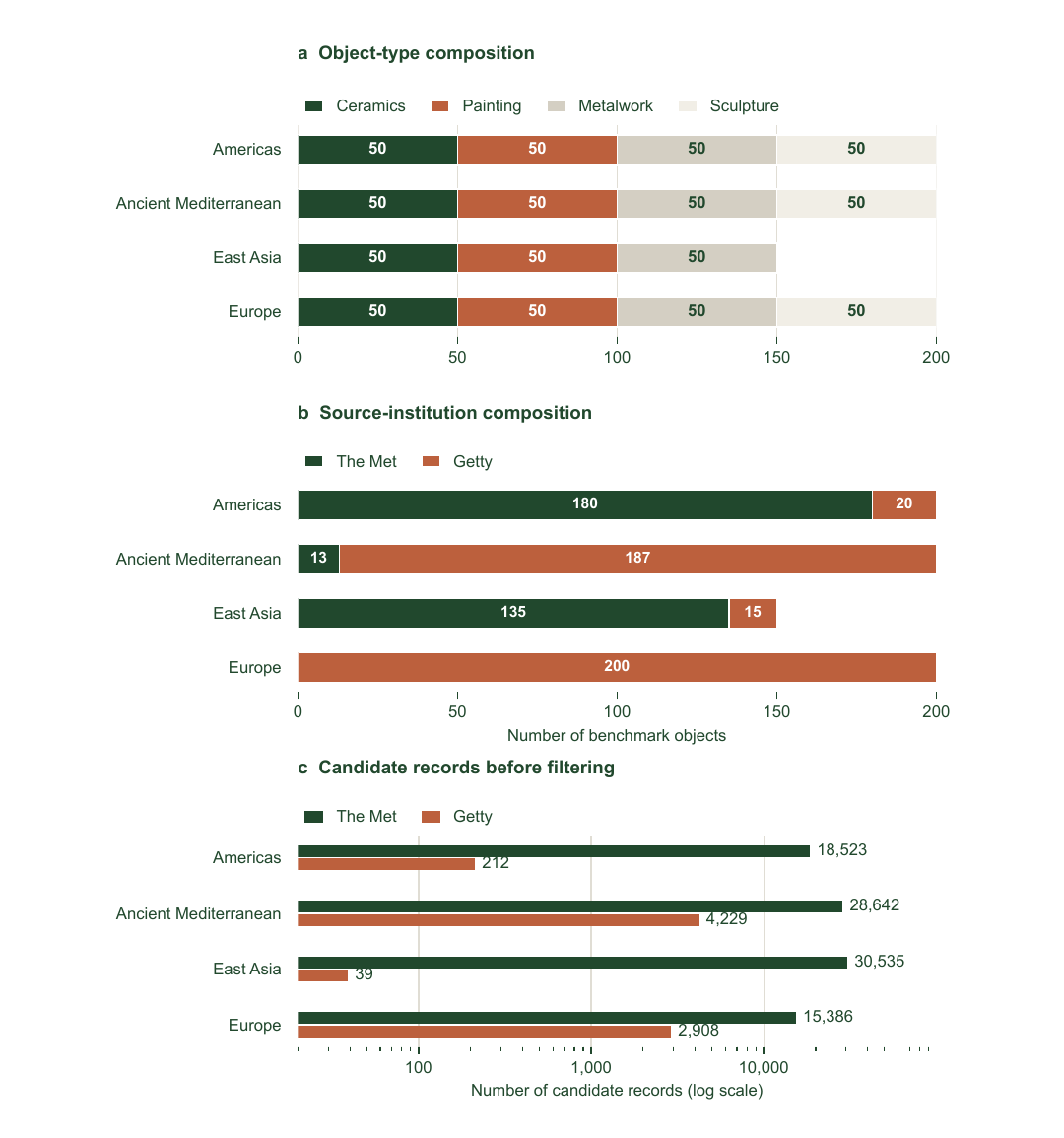}
    \caption{Benchmark composition: (a) object types; (b) source institutions; (c) pre-filtering candidates (log scale). There is no East Asian sculpture cell.}
    \label{fig:dataset-distribution}
\end{figure}

After filtering for metadata, images, category consistency, and cross-regional coverage, one annotator selected 50 objects per available region-type cell; a second audited a stratified 20\% sample ($n=150$), and disagreements were corrected. Categories follow museum vocabularies~\cite{harpring2010controlled,baca2016metadata,doerr2003cidoc}, with bronzes included in metalwork~\cite{moc2001grading}. The diagnostic, evaluation-only set contains 750 objects in 15 cells: 200 each from the Americas, Ancient Mediterranean, and Europe across four types, and 150 from East Asia across three types because East Asian sculpture lacked sufficient coverage (Figure~\ref{fig:dataset-distribution}; Appendix~\ref{appendix:dataset-statistics}). These curator-written records serve as the reference attributions. 

\subsection{Evaluation Protocol}
\label{sec:evaluation}

Each LLM judge scores the five fields independently: 1 indicates semantic compatibility with an applicable reference label, and 0 incompatibility. Inputs contain predictions, reference-label arrays, and museum descriptions, with human decisions, model identities, and sampling strata withheld. Appendix~\ref{appendix:prompt} provides the prompt.

The human calibration sample has five outputs per region-type cell and model (675). Two annotators independently review each output, and a third who did not initially label that group independently adjudicates disagreements. Four candidate LLM judges are compared: GPT-4.1-mini (GPT-J), DeepSeek-V4.1-Flash (DS-J), Gemini-3.5-Flash-Lite (Gem-J), and Qwen3-30B-A3B-Instruct-2507 (Qwen-J) (Table~\ref{tab:candidate-judge-models}). Calibration compares all four against the same eligible human-labelled fields using agreement, binary $F_1$ (positive label 1), and Cohen's $\kappa$. The highest pooled $\kappa$ selects the full-benchmark judge (Section~\ref{sec:human-evaluation}).

For empty references, automatic scores are 1 for an unknown or uncertain answer and 0 for a definite answer, following cataloguing practice that marks hypotheses as tentative when evidence is lacking. Calibration retains the 190 binary human labels on these fields and excludes unlabelled ones. We report nonempty-reference accuracy and empty-reference uncertainty handling separately; Mean, Ctx, and Acc combine the two rules (Table~\ref{tab:reference-split}). We report the distribution of the compatible-field count $s_i$. Ctx excludes title, which names an object type for decorative objects but a specific work for paintings and has the lowest human field-level $\kappa$ (0.634; Table~\ref{tab:judge-agreement}). Valid outputs parse into the five-field schema; invalid outputs are excluded from every accuracy denominator.

\section{Experiment}
We evaluate the nine open-weight and proprietary endpoints in Table~\ref{tab:vlm-overview} on $750\times9=6{,}750$ outputs, without demonstrations. Predictions use the VLM prompt in Appendix~\ref{appendix:prompt}. The selected judge DS-J uses temperature 0, a 1,200-token limit, and disabled thinking; format failures are retried up to four times per pass, and successful decisions are cached. Appendix~\ref{appendix:prompt} contains prompts and output contracts.

\subsection{Human Evaluation}\label{sec:human-evaluation}
Appendices~\ref{appendix:annotator-background} and~\ref{appendix:annotation-guideline} document the annotation workbooks, protocol, and guideline. The completed annotations contain 3,187 eligible human field labels. Table~\ref{tab:judge-agreement} reports reliability before adjudication.

\begin{table}[!t]
\caption{Human reliability before adjudication (agreement and $F_1$ in \%; positive label 1).}
\label{tab:judge-agreement}
\centering
\setlength{\tabcolsep}{3pt}\fontsize{9}{11}\selectfont
\begin{tabular*}{\columnwidth}{@{\extracolsep{\fill}}lrrrr@{}}
\toprule
Field & Valid labels & Agreement (\%) & $F_1$ (\%) & $\kappa$ \\
\midrule
Title & 666 & 81.68 & 81.17 & 0.634 \\
Culture & 665 & 94.59 & 94.67 & 0.892 \\
Period & 591 & 86.46 & 87.34 & 0.729 \\
Origin & 566 & 89.93 & 89.42 & 0.798 \\
Creator & 557 & 89.41 & 84.68 & 0.766 \\
\midrule
Overall & 3,045 & 88.37 & 87.71 & 0.767 \\
\bottomrule
\end{tabular*}

\end{table}

\begin{figure}[!htbp]
\centering
\includegraphics[width=\columnwidth,height=0.70\textheight,keepaspectratio]{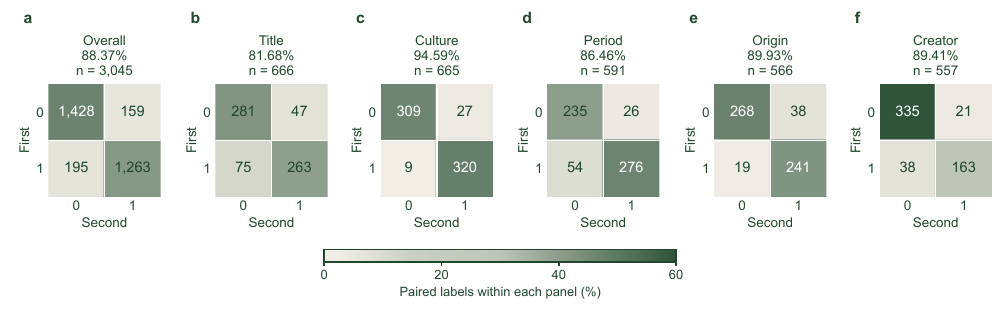}
\caption{Inter-annotator confusion matrices before adjudication (rows: first annotator; 0/1: incompatible/compatible). Colour: \% of pairs per panel.}
\label{fig:human-confusion}
\end{figure}

Before adjudication, human annotations achieve 88.37\% agreement, 87.71\% binary $F_1$, and $\kappa=0.767$ (Figure~\ref{fig:human-confusion}). Agreement is highest for culture (94.59\%) and lowest for title (81.68\%), where concise museum names and LLM titles with added visual descriptions lead to disagreements.

Table~\ref{tab:judge-selection} reports candidate agreement with the human reference; Table~\ref{tab:human-judge-accuracy} in Appendix~\ref{appendix:additional-results} gives per-model compatible-field rates.

\begin{table}[!htbp]
\caption{Candidate-judge agreement with final human labels (agreement and $F_1$ in \%; positive label 1).}
\label{tab:judge-selection}
\centering
\setlength{\tabcolsep}{3pt}\fontsize{8}{9.5}\selectfont
\begin{tabular}{lrrr}
\toprule
Candidate Judge & Agreement (\%) & $F_1$ (\%) & $\kappa$ \\
\midrule
GPT-J & 87.79 & 86.01 & 0.753 \\
DS-J & 89.46 & 88.19 & 0.787 \\
Gem-J & 85.97 & 84.31 & 0.717 \\
Qwen-J & 84.37 & 81.97 & 0.683 \\
\bottomrule
\end{tabular}
\end{table}

\begin{figure}[!htbp]
\centering
\includegraphics[width=\columnwidth,height=0.70\textheight,keepaspectratio]{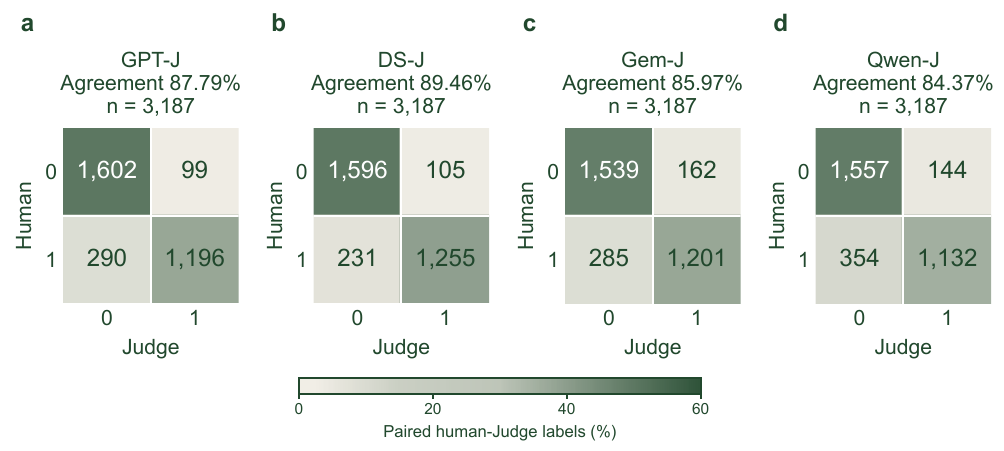}
\caption{Candidate-judge confusion matrices on 3,187 human labels (rows: human; 0/1: incompatible/compatible). Colour: \% of all pairs.}
\label{fig:candidate-judge-confusion}
\end{figure}

\begin{figure}[!htbp]
\centering
\includegraphics[width=\columnwidth,height=0.70\textheight,keepaspectratio]{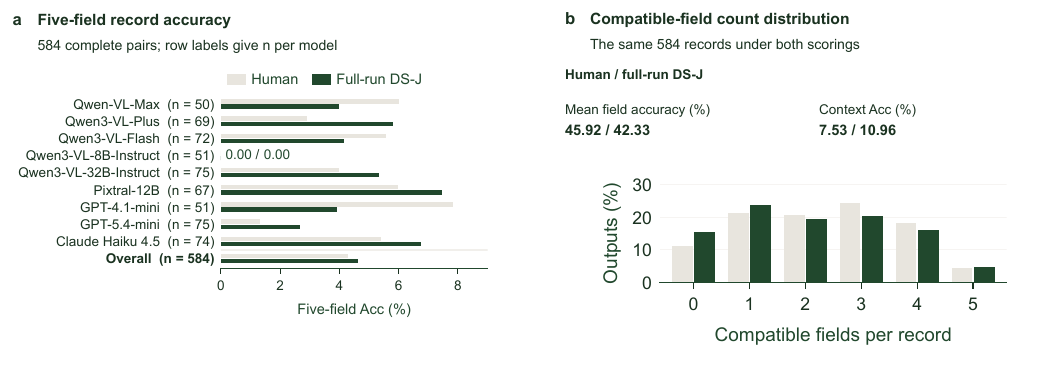}
\caption{Human-scored subset: (a) per-model five-field Acc; (b) compatible-field count distributions, with pooled Mean and Ctx annotated. Both scorings use the same 584 complete records; Mean uses their 2,920 fields. DS-J uses the full run. Values are percentages.}
\label{fig:matched-subset-metrics}
\end{figure}

DS-J, selected for the full run, has the highest candidate-stage $\kappa$ (0.787), with 89.46\% agreement and 88.19\% $F_1$; GPT-J follows ($\kappa=0.753$) (Figure~\ref{fig:candidate-judge-confusion}). Full-run agreement with the same 3,187 final human labels is 88.05\%, with $\kappa=0.759$ (95\% CI [0.729, 0.786]). The two runs differ on 145 fields (4.55\%; Appendix~\ref{sec:dsj-run-stability}).
\paragraph{Human-scored subset.}
Of 675 reviewed outputs, 19 are invalid and 584 have all five human labels. Table~\ref{tab:matched-subset-metrics} and Figure~\ref{fig:matched-subset-metrics} compare human-scored and full-run DS-J accuracy and compatible-field distributions on these same records. Human and DS-J Acc are close (4.28\% vs.\ 4.62\%). Missing labels are excluded without imputing zero. Per-model field agreement, candidate/full-run differences, and bootstrap ranking instability are detailed in Appendix~\ref{appendix:additional-results}.

\begin{table}[!htbp]
\caption{Human-scored and full-run DS-J results on the same 584 complete records from 675 sampled outputs (19 invalid). $n_c$: paired records. Mean uses 2,920 fields; Ctx excludes title. The distribution concerns this human-labelled subset, not the 6,542-output benchmark. Shares are rounded.}
\label{tab:matched-subset-metrics}
\centering
\fontsize{8}{9.5}\selectfont
\setlength{\tabcolsep}{2pt}
\textbf{(a) Record accuracy (\%)}\\[2pt]
\begin{tabular*}{\columnwidth}{@{\extracolsep{\fill}}>{\scriptsize}lrrr@{}}
\toprule
Model & $n_c$ & Human & Full-run DS-J \\
\midrule
Qwen-VL-Max & 50 & 6.00 & 4.00 \\
Qwen3-VL-Plus & 69 & 2.90 & 5.80 \\
Qwen3-VL-Flash & 72 & 5.56 & 4.17 \\
\makecell[l]{Qwen3-VL-8B-\\Instruct} & 51 & 0.00 & 0.00 \\
\makecell[l]{Qwen3-VL-32B-\\Instruct} & 75 & 4.00 & 5.33 \\
Pixtral-12B & 67 & 5.97 & 7.46 \\
GPT-4.1-mini & 51 & 7.84 & 3.92 \\
GPT-5.4-mini & 75 & 1.33 & 2.67 \\
Claude Haiku 4.5 & 74 & 5.41 & 6.76 \\
\midrule
Overall & 584 & 4.28 (25/584) & 4.62 (27/584) \\
\bottomrule
\end{tabular*}
\par\smallskip
\textbf{(b) Pooled field and context accuracy}\\[2pt]
\begin{tabular*}{\columnwidth}{@{\extracolsep{\fill}}lcc@{}}
\toprule
Metric & Human & Full-run DS-J \\
\midrule
Mean (\%) & 45.92 & 42.33 \\
Context Acc (\%) & 7.53 (44/584) & 10.96 (64/584) \\
\bottomrule
\end{tabular*}
\par\smallskip\textbf{(c) Human-labelled subset: compatible-field counts}\\[2pt]
\begin{tabular*}{\columnwidth}{@{\extracolsep{\fill}}lrrrrrr@{}}
\toprule
$s_i$ & 0 & 1 & 2 & 3 & 4 & 5 \\
\midrule
Human (\%) & 11.30 & 21.23 & 20.72 & 24.32 & 18.15 & 4.28 \\
Full-run DS-J (\%) & 15.58 & 23.80 & 19.35 & 20.55 & 16.10 & 4.62 \\
\bottomrule
\end{tabular*}
\end{table}

\subsection{Automatic Evaluation Results}
\begin{table}[!tp]
\caption{Full-run DS-J results (\% of valid outputs). Model Mean/Ctx/Acc cells include 95\% object-bootstrap CIs (10,000 resamples); the pooled row gives point estimates. Inv.: invalid; Mean: mean field accuracy; Ctx: four contextual fields compatible; Acc: all five fields compatible. ---: not reported. Values are point estimates, with no rank highlighting.}
\label{tab:overall}
\centering
\fontsize{8}{9.5}\selectfont
\setlength{\tabcolsep}{1.6pt}
\begin{tabular*}{\textwidth}{@{\extracolsep{\fill}}>{\scriptsize}lrrrrrrrrrr@{}}
\toprule
Model & Inv. & Valid $n$ & Mean & Ctx & Acc & Title & Cul. & Per. & Ori. & Cre. \\
\midrule
Qwen-VL-Max & 38 & 712 & 46.57\,{\scriptsize[44.50, 48.68]} & 11.66\,{\scriptsize[9.32, 14.11]} & 5.90\,{\scriptsize[4.23, 7.69]} & 35.11 & 57.72 & 55.90 & 49.58 & 34.55   \\
Qwen3-VL-Plus & 39 & 711 & 42.33\,{\scriptsize[40.08, 44.59]} & 10.97\,{\scriptsize[8.77, 13.32]} & 5.63\,{\scriptsize[4.02, 7.38]} & 32.07 & 62.31 & 42.62 & 47.96 & 26.72   \\
Qwen3-VL-Flash & 39 & 711 & 49.82\,{\scriptsize[47.69, 51.86]} & 17.02\,{\scriptsize[14.33, 19.78]} & 7.03\,{\scriptsize[5.22, 8.95]} & 32.49 & 65.12 & 55.98 & 55.41 & 40.08   \\
Qwen3-VL-8B-Instruct & 0 & 750 & 36.16\,{\scriptsize[34.19, 38.11]} & 7.20\,{\scriptsize[5.47, 9.07]} & 2.13\,{\scriptsize[1.20, 3.20]} & 22.13 & 43.73 & 39.60 & 42.00 & 33.33   \\
Qwen3-VL-32B-Instruct & 0 & 750 & 40.21\,{\scriptsize[38.08, 42.37]} & 11.20\,{\scriptsize[9.07, 13.47]} & 4.80\,{\scriptsize[3.33, 6.40]} & 25.73 & 50.40 & 45.20 & 44.93 & 34.80   \\
Pixtral-12B & 74 & 676 & 41.51\,{\scriptsize[39.27, 43.78]} & 14.05\,{\scriptsize[11.49, 16.76]} & 5.77\,{\scriptsize[4.04, 7.57]} & 28.40 & 46.75 & 45.56 & 45.86 & 40.98   \\
GPT-4.1-mini & 0 & 750 & 48.61\,{\scriptsize[46.51, 50.75]} & 15.47\,{\scriptsize[12.93, 18.13]} & 7.60\,{\scriptsize[5.73, 9.60]} & 36.67 & 60.27 & 58.93 & 50.80 & 36.40   \\
GPT-5.4-mini & 0 & 750 & 46.88\,{\scriptsize[44.69, 49.09]} & 16.53\,{\scriptsize[14.00, 19.20]} & 7.87\,{\scriptsize[6.00, 9.87]} & 34.93 & 59.07 & 50.27 & 51.07 & 39.07   \\
Claude Haiku 4.5 & 18 & 732 & 44.10\,{\scriptsize[41.92, 46.33]} & 15.71\,{\scriptsize[13.17, 18.38]} & 7.38\,{\scriptsize[5.59, 9.32]} & 29.78 & 52.87 & 49.45 & 46.04 & 42.35   \\
\midrule
Overall & 208 & 6,542 & 44.01 & 13.30 & 6.01 & \multicolumn{5}{c}{---} \\
\bottomrule
\end{tabular*}
\end{table}
\begin{table}[!t]
\caption{Reference-aware field scores and independent-field baselines (\%). Recorded/Empty: scores on nonempty/empty reference arrays; $n_R,n_E$: eligible fields. Recorded includes explicit unknown-maker labels. Ind.: product of five marginal field scores; Cond.: product within reference-availability patterns, weighted by valid-output frequency. $\Delta$: Acc minus Cond. in percentage points, with paired object-bootstrap 95\% CI (10,000 resamples).}
\label{tab:reference-split}
\label{tab:independence}
\centering\fontsize{8}{10}\selectfont
\setlength{\tabcolsep}{2pt}
\begin{tabular*}{\textwidth}{@{\extracolsep{\fill}}lrrrrrrrr@{}}
\toprule
Model & $n_R$ & Recorded & $n_E$ & Empty & Acc & Ind. & Cond. & $\Delta$ [95\% CI] \\
\midrule
Qwen-VL-Max & 3,249 & 44.78 & 311 & 65.27 & 5.90 & 1.94 & 2.39 & +3.51 [+2.11, +4.94] \\
Qwen3-VL-Plus & 3,244 & 41.06 & 311 & 55.63 & 5.63 & 1.09 & 1.84 & +3.79 [+2.29, +5.21] \\
Qwen3-VL-Flash & 3,244 & 47.66 & 311 & 72.35 & 7.03 & 2.63 & 3.12 & +3.92 [+2.43, +5.39] \\
Qwen3-VL-8B & 3,432 & 33.51 & 318 & 64.78 & 2.13 & 0.54 & 0.57 & +1.57 [+0.63, +2.54] \\
Qwen3-VL-32B & 3,432 & 37.47 & 318 & 69.81 & 4.80 & 0.92 & 1.04 & +3.76 [+2.45, +5.16] \\
Pixtral-12B & 3,102 & 39.36 & 278 & 65.47 & 5.77 & 1.14 & 1.65 & +4.12 [+2.62, +5.57] \\
GPT-4.1-mini & 3,432 & 47.29 & 318 & 62.89 & 7.60 & 2.41 & 2.73 & +4.87 [+3.24, +6.50] \\
GPT-5.4-mini & 3,432 & 45.54 & 318 & 61.32 & 7.87 & 2.07 & 2.47 & +5.40 [+3.72, +6.96] \\
Claude Haiku 4.5 & 3,352 & 41.89 & 308 & 68.18 & 7.38 & 1.52 & 1.96 & +5.42 [+3.85, +6.96] \\
\midrule
Pooled & 29,919 & 42.05 & 2,791 & 65.07 & 6.01 & \multicolumn{3}{c}{---} \\
\bottomrule
\end{tabular*}
\end{table}

DS-J scores the 6,542 valid outputs; the 208 invalid outputs are reported separately. Qwen API invalid outputs result from provider-side safety filtering; Pixtral-12B failures return no parseable record. Counting them as incorrect lowers Acc from 5.77\% to 5.20\% for Pixtral-12B and from 5.90\% to 5.60\% for Qwen-VL-Max, reversing only Pixtral-12B and Qwen3-VL-Plus in the Acc ranking.

\begin{figure}[!htbp]
\centering
\includegraphics[width=0.60\linewidth,height=0.70\textheight,keepaspectratio]{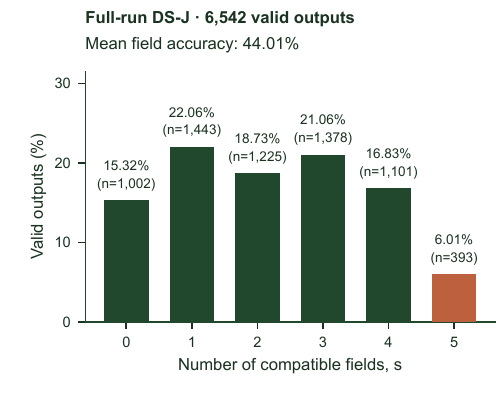}
\caption{Compatible-field count distribution for 6,542 valid full-run DS-J outputs. Labels give percentages and counts; the all-five-compatible bin contains 393 records. The pooled Mean is 44.01\%.}
\label{fig:compatible-field-distribution}
\end{figure}

\paragraph{Compatible-field distribution.}
Figure~\ref{fig:compatible-field-distribution} reports all six counts, with pooled Mean 44.01\%; Table~\ref{tab:compatible-field-distribution} in Appendix~\ref{appendix:additional-results} gives the exact values. Most outputs have one to four compatible fields; 6.01\% have all five.

\begin{figure}[!t]
\centering
\includegraphics[width=\columnwidth,height=0.70\textheight,keepaspectratio]{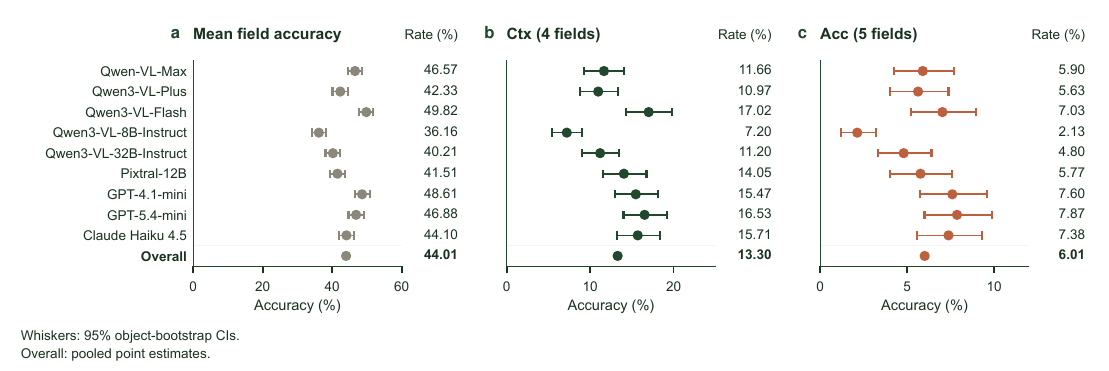}
\captionsetup{skip=3pt}
\caption{Full-run DS-J comparison of Mean, four-field Ctx, and five-field Acc reported as percentages. Whiskers are model-level 95\% object-bootstrap CIs (10,000 resamples); the pooled row shows point estimates. The all-four-field Ctx rate is 13.30\%, compared with 6.01\% for all-five-field Acc.}
\label{fig:field-record-results}
\end{figure}

\paragraph{Endpoint comparison.}
Qwen3-VL-Flash has the highest point estimates for Mean (49.82\%) and Ctx (17.02\%), and GPT-5.4-mini for Acc (7.87\%; Table~\ref{tab:overall}). Overlapping intervals and the weak human-judge Acc rank correlation ($\rho=0.283$ on the human subset; Appendix~\ref{appendix:additional-results}) limit fine-grained ranking. Excluding title raises the pooled all-compatible rate from 6.01\% to 13.30\%. The independent-field baseline below tests whether these complete records are more than chance combinations of correct fields.

\paragraph{Recorded and empty references.}
Table~\ref{tab:reference-split} separates nonempty-reference accuracy (42.05\% over 29,919 fields) from empty-reference uncertainty handling (65.07\% over 2,791 fields). Their field-count-weighted average is the original Mean, 44.01\%. Nonempty references include explicit unknown-maker labels; creator scores on named-reference objects are further separated in Section~\ref{sec:title-ablation}. Appendix~\ref{appendix:reference-diagnostics} gives per-field scores and denominators.

\paragraph{Independent-field baseline.}
If field decisions were independent with the observed marginal scores $A_f$, their all-correct probability would be $\prod_f A_f$. We also compute this product within each reference-availability pattern and average by valid-output frequency (Appendix~\ref{appendix:metric-definitions}). Observed Acc (2.13--7.87\%) exceeds both the unconditional expectation (0.54--2.63\%) and the conditioned expectation (0.57--3.12\%) for every model. Observed-minus-conditioned differences are +1.57 to +5.42 points, with paired bootstrap intervals above zero (Table~\ref{tab:independence}). Correct fields thus cluster on some objects, which models attribute as a whole, while for most objects they recover some fields and miss others. Ctx shows the same pattern (Appendix~\ref{appendix:reference-diagnostics}).

Culture scores highest for every endpoint; creator is the lowest contextual field, and title scores below creator for six of nine endpoints (Table~\ref{tab:overall}). East Asia has the highest regional Acc for eight endpoints. The lowest is Europe for five, the Ancient Mediterranean for three, and the Americas for GPT-5.4-mini (Table~\ref{tab:regional-results}).

\begin{table}[!t]
\caption{Qualitative cases of metadata incompatibility.}
\label{tab:error-case-summary}
\centering
\fontsize{8}{9.5}\selectfont
\setlength{\tabcolsep}{3pt}
\begin{tabular*}{\textwidth}{@{\extracolsep{\fill}}>{\centering\arraybackslash}m{.07\textwidth}>{\raggedright\arraybackslash}m{.26\textwidth}>{\raggedright\arraybackslash}m{.35\textwidth}>{\raggedright\arraybackslash}m{.25\textwidth}@{}}
\toprule
& \textbf{Case and recorded reference} & \textbf{Representative model prediction} & \textbf{Diagnostic observation} \\
\midrule
\includegraphics[width=\linewidth,height=28pt,keepaspectratio]{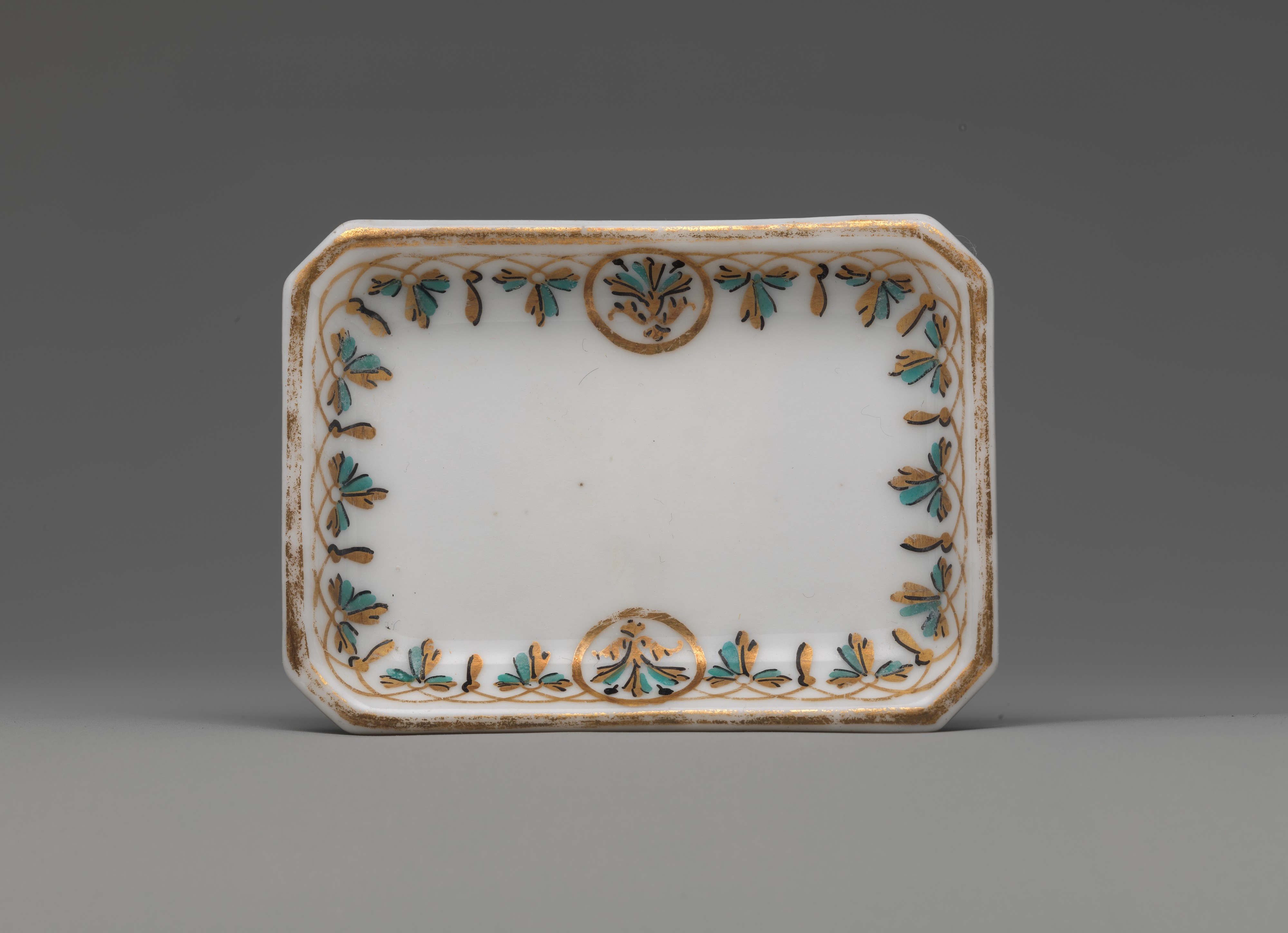} & \textbf{A. Butter Pat}: American, 1885; Union Porcelain Works.
& GPT-4.1-mini: European, France, eighteenth century.
& The predicted culture and period conflict with the museum record. \\[1pt]
\includegraphics[width=\linewidth,height=28pt,keepaspectratio]{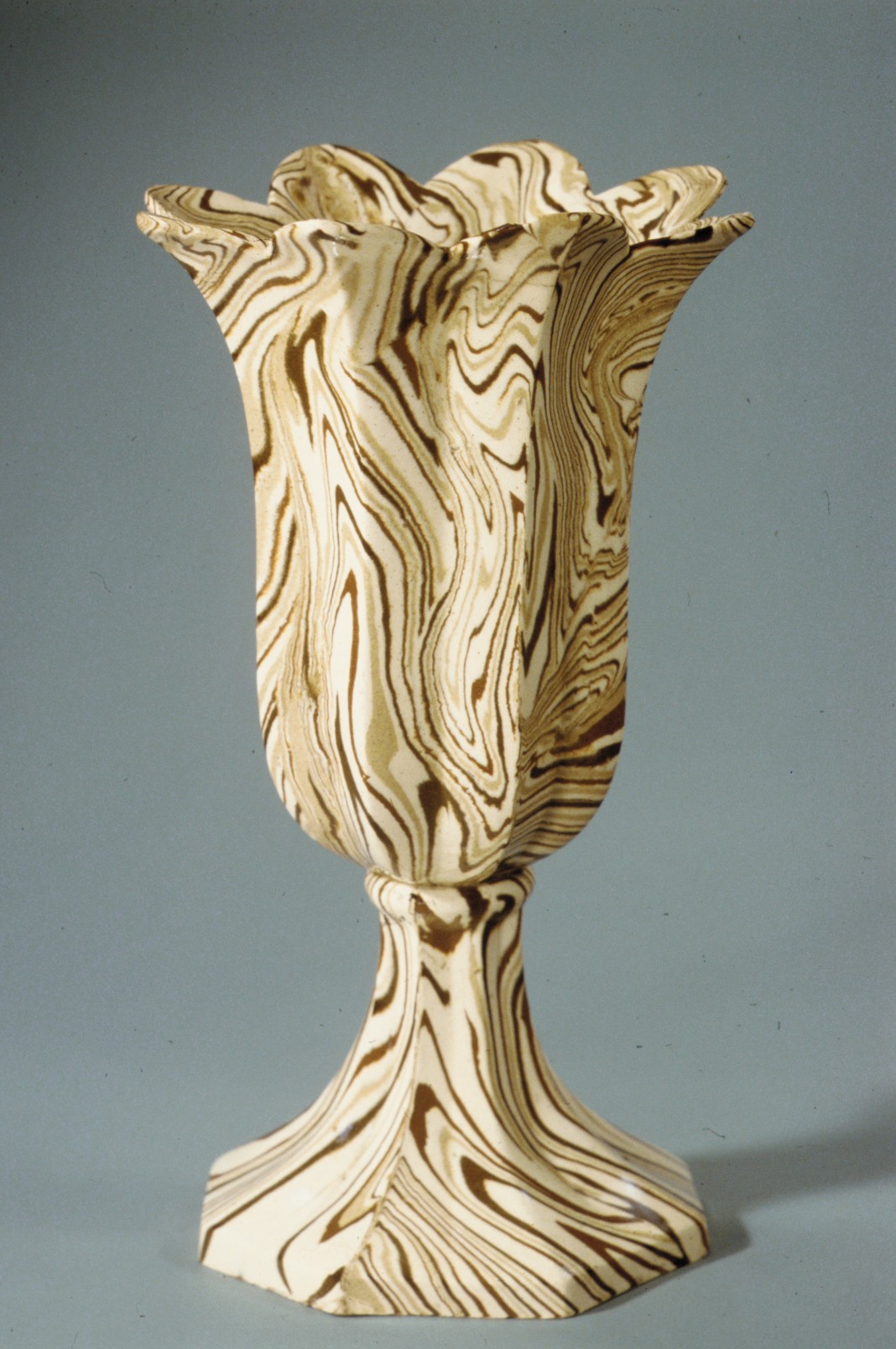} & \textbf{B. Celery vase}: American, 1849--1858; United States Pottery Company.
& GPT-4.1-mini: Dutch, Delftware workshop.
& Cultural misattribution accompanies an incompatible maker attribution. \\[1pt]
\includegraphics[width=\linewidth,height=28pt,keepaspectratio]{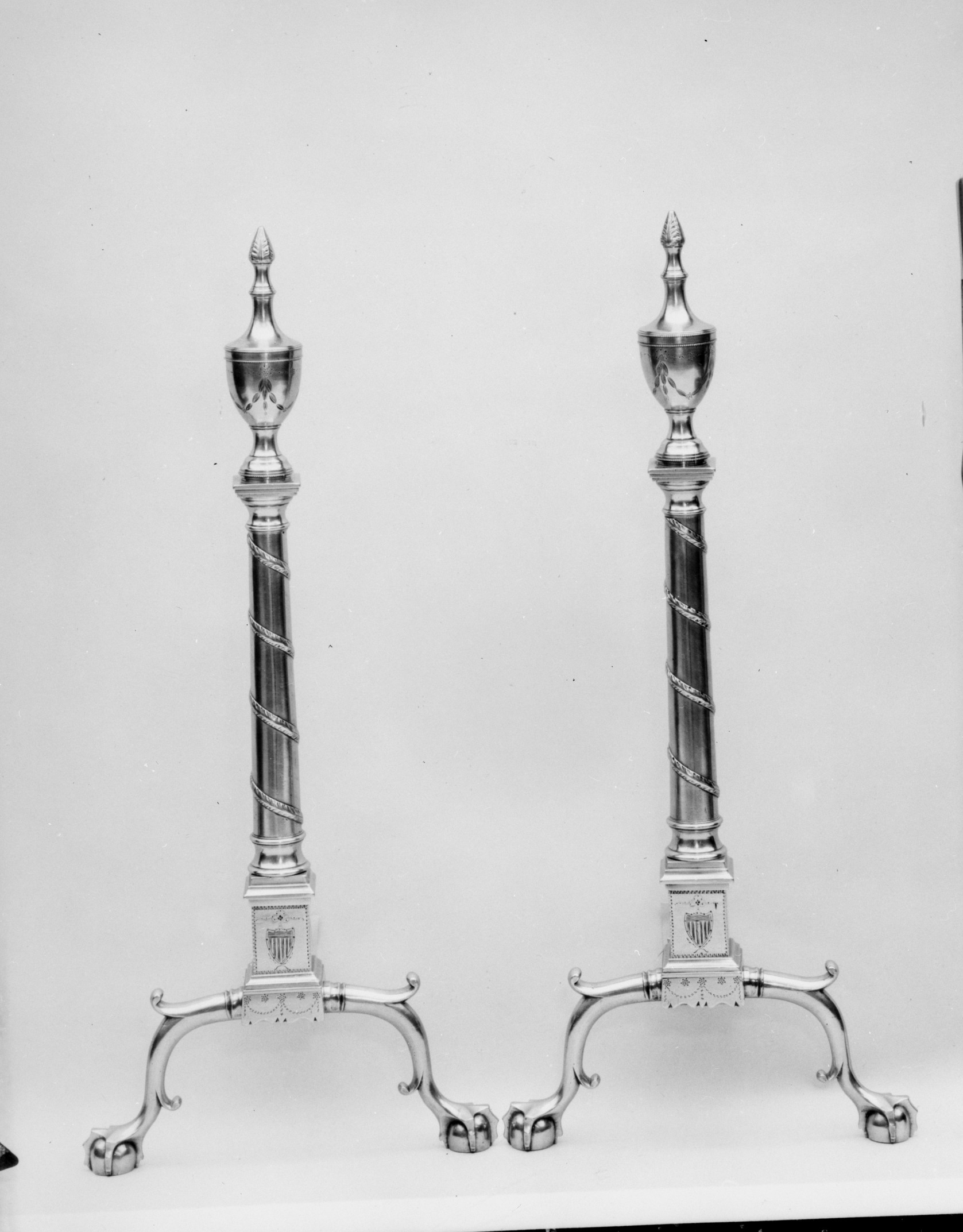} & \textbf{C. Andiron}: American, 1795--1810.
& Pixtral-12B: \emph{Pair of Empire Style Andirons}; Western; early nineteenth century; France.
& Title and period are compatible; culture, origin, and creator are incompatible. \\[1pt]
\includegraphics[width=\linewidth,height=28pt,keepaspectratio]{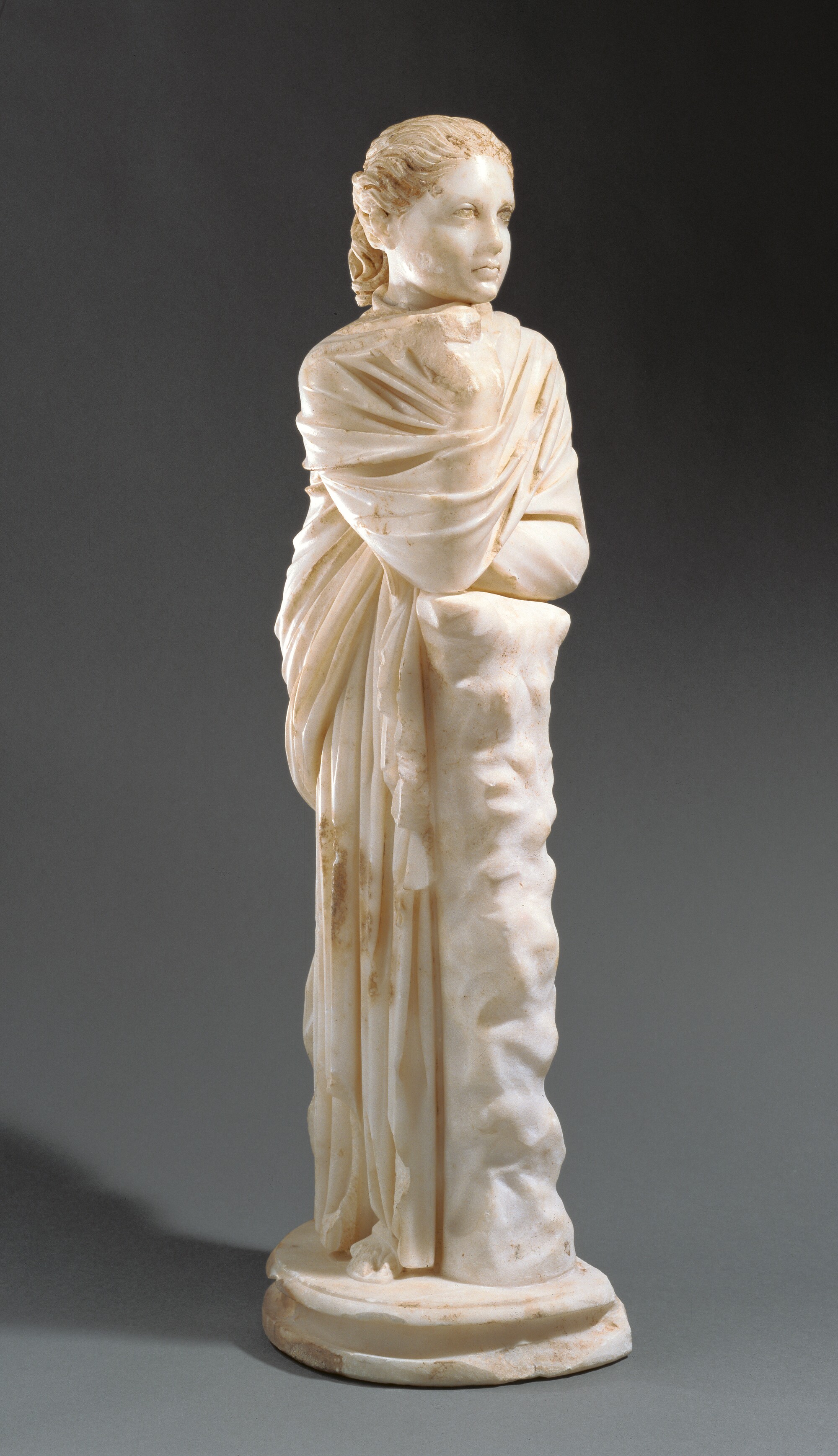} & \textbf{D. Statue of a Muse}: Roman.
& GPT-5.4-mini: \emph{Funerary statue of a young woman}; Roman; Eastern Mediterranean. Pixtral-12B: \emph{Standing Female Figure}; Roman.
& Roman culture matches; the titles differ from the recorded Muse identification. \\[1pt]
\includegraphics[width=\linewidth,height=28pt,keepaspectratio]{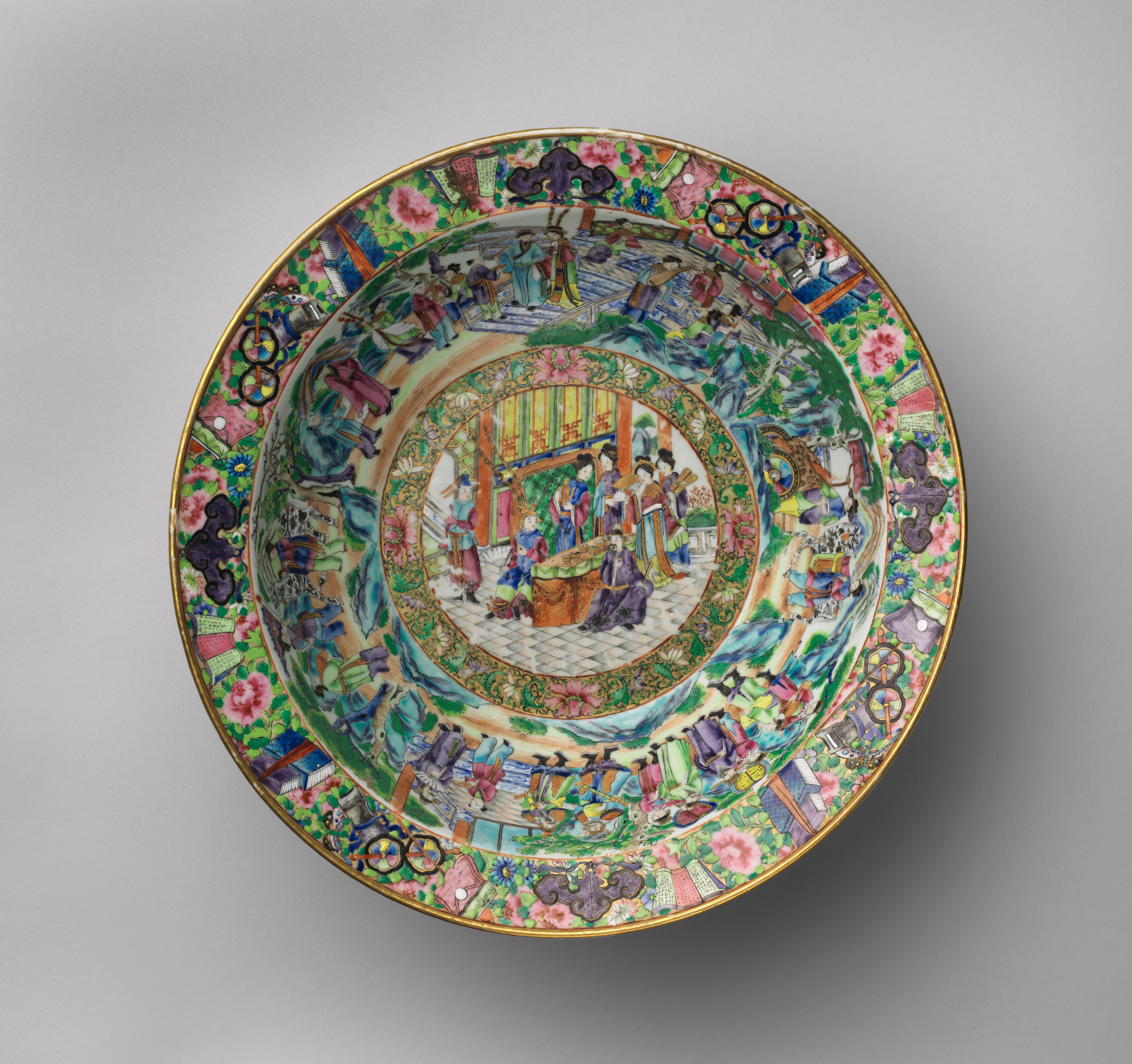} & \textbf{E. Basin}: Chinese, 1825--1845.
& Qwen3-VL-Plus: Guangxu period (1875--1908), with Eight Immortals.
& Chinese culture matches; the predicted reign differs from the recorded date. \\[1pt]
\includegraphics[width=\linewidth,height=28pt,keepaspectratio]{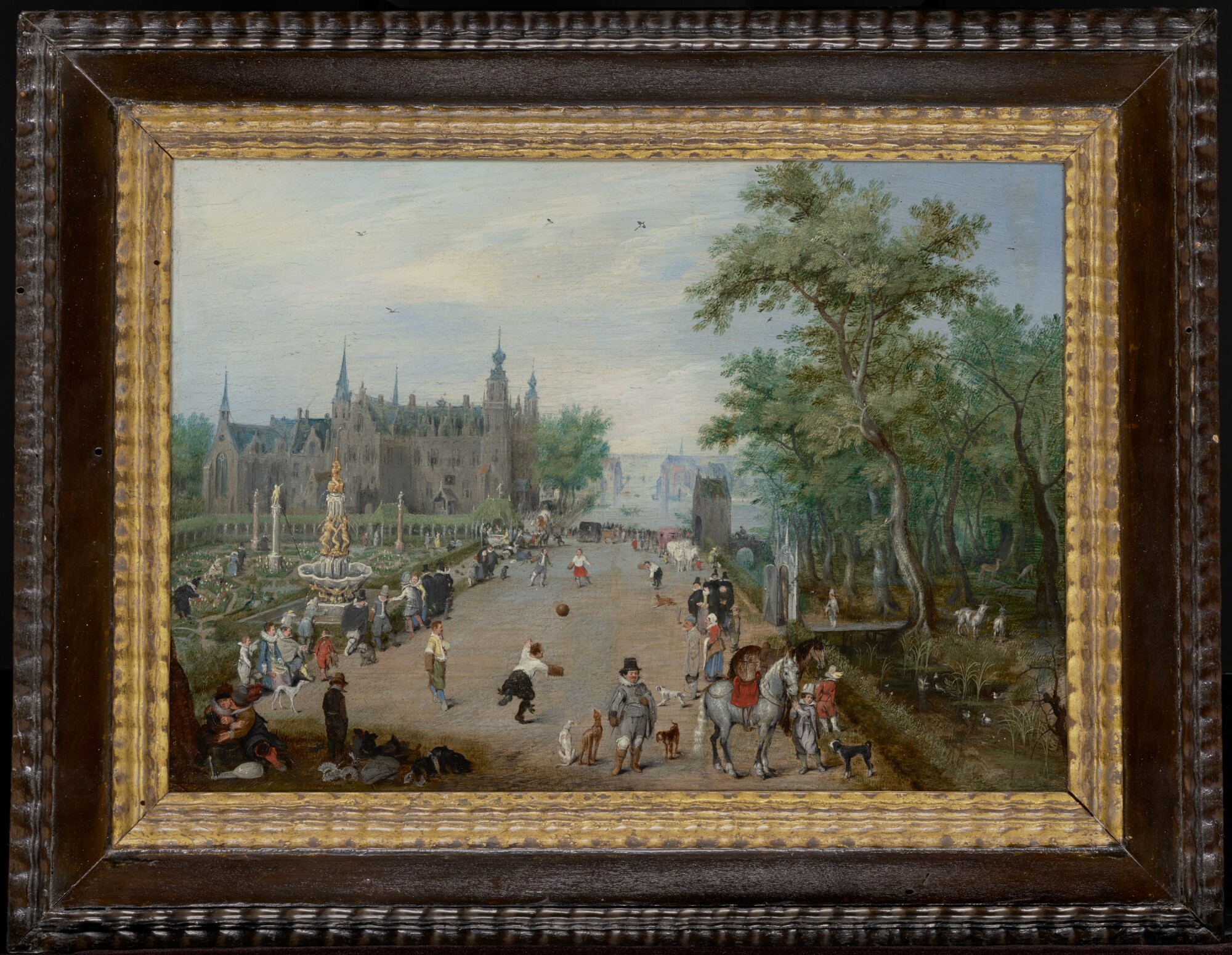} & \textbf{F. A Ball Game Before a Country Palace}: Dutch, seventeenth century; Adriaen van de Venne.
& Pixtral-12B: \emph{View of the Courtyard of the Amsterdam City Hall}; Golden Age; possibly a follower of Pieter Saenredam.
& The culture label is compatible, but the title and creator are incompatible. \\
\bottomrule
\end{tabular*}
\end{table}

\subsection{Museum-and-Title-Only Ablation}
\label{sec:title-ablation}
\begin{table}[H]
\caption{Museum-and-title-only ablation: DS-J field accuracy (\%). Images: Table~\ref{tab:overall}.}
\label{tab:title-ablation}
\centering
\fontsize{8}{10}\selectfont
\setlength{\tabcolsep}{3pt}
\begin{tabular*}{\columnwidth}{@{\extracolsep{\fill}}llrrrrr@{}}
\toprule
Model & Input & Cul. & Per. & Ori. & Cre. & Ctx \\
\midrule
GPT-4.1-mini & Images & 60.3 & 58.9 & 50.8 & 36.4 & 15.5 \\
 & Museum + title & 31.5 & 26.7 & 36.1 & 43.7 & 7.1 \\
\midrule
GPT-5.4-mini & Images & 59.1 & 50.3 & 51.1 & 39.1 & 16.5 \\
 & Museum + title & 36.9 & 32.9 & 38.5 & 44.1 & 8.9 \\
\bottomrule
\end{tabular*}
\end{table}

For GPT-4.1-mini and GPT-5.4-mini, we compare image-only predictions with predictions of culture, period, origin, and creator from the collection name (Met or Getty) and museum title for the same 750 objects. DS-J uses the same references in both conditions (Table~\ref{tab:title-ablation}). Museum-and-title-only input scores 12.6--32.2 points below image input on culture, period, and origin, and 8.4 and 7.6 points below on Ctx; all object-level bootstrap 95\% CIs exclude zero. Without images, both models abstain on 50--61\% of these fields. Titles such as \emph{Wakizashi} or \emph{Kylix} supply culture-specific cues, whereas \emph{Plate} and \emph{Bowl} supply few.

Creator scores rise by 7.3 and 5.1 points as both models abstain on about 97\% of creators. The gain comes from the 201 explicit unknown-maker references and 126 empty arrays, while scores on the 423 objects with at least one named-creator attribution fall from 15.6\% to 1.2\% and from 13.5\% to 2.4\%. GPT-4.1-mini gains 100 matches on explicit unknown labels and 16 uncertainty-handling successes on empty references, but loses 61 matches on named-reference objects; GPT-5.4-mini gains 67 and 18 while losing 47. Aggregate creator scores therefore mask lower scores on named-reference objects (Appendix~\ref{appendix:reference-diagnostics}).

\noindent\textbf{Specific-object identification.}
Given each original photograph, GPT-4.1-mini claims to identify 16 of the 750 objects, and only 3 of these claims are verified against official catalogues (Table~\ref{tab:object-identification}).

\begin{table}[!htbp]
\setlength{\abovecaptionskip}{2pt}
\caption{GPT-4.1-mini closed-book identification ($N=750$).}
\label{tab:object-identification}
\centering
\fontsize{8}{9.5}\selectfont
\setlength{\tabcolsep}{4pt}
\begin{tabular*}{\columnwidth}{@{\extracolsep{\fill}}lrr@{}}
\toprule
Identification & Count & \% of 750 \\
\midrule
Claimed & 16 & 2.13 \\
Locally verified & 3 & 0.40 \\
\bottomrule
\end{tabular*}
\end{table}

\subsection{Error Analysis}
\label{sec:errors}
Table~\ref{tab:error-case-summary} and Appendix~\ref{appendix:error-analysis} illustrate four error types.

\par\smallskip\noindent\textbf{Cross-cultural misattribution.}
Models assign objects to cultural traditions that differ from the museum record. The American \emph{Butter Pat} is labelled European by GPT-4.1-mini, Japanese by Qwen-VL-Max, and Chinese by Pixtral-12B (Case~A). For the American \emph{Celery vase}, GPT-4.1-mini predicts Dutch culture and a Delftware workshop, replacing both the recorded culture and maker (Case~B). In both cases the attribution follows the style of the object rather than the American makers recorded by the museum.

\par\smallskip\noindent\textbf{Object-type and identity confusion.}
Recognising visible form does not ensure the correct object type or identity. GPT-4.1-mini calls \emph{Compote} an \emph{Allegorical Plate with Justice}; DS-J rejects the title despite accepting its other four fields. For \emph{Statue of a Muse}, Pixtral-12B gives \emph{Standing Female Figure}, missing the Muse identification (Case~D).

\par\smallskip\noindent\textbf{Temporal misattribution.}
Incorrect dates occur both with and without cultural misattribution. GPT-4.1-mini shifts the American \emph{Butter Pat} from 1885 to the eighteenth century (Case~A). For the Chinese \emph{Basin}, Qwen3-VL-Plus predicts the Guangxu reign (1875--1908) and Claude Haiku 4.5 the Kangxi period (1662--1722), although both recover Chinese culture; neither period matches the recorded 1825--1845 date (Case~E).

\par\smallskip\noindent\textbf{Creator agreement with other field errors.}
For \emph{Compote}, Qwen3-VL-Flash correctly names Union Porcelain Works but predicts Neoclassical European culture, an earlier period, and France or Germany; DS-J rejects all four other fields. Across valid outputs, 1,992 of 2,385 positive creator decisions (83.52\%, including unknown and empty references) coexist with another incompatible field. Knowing the maker therefore does not bind the other fields to the same object.

Most of these errors follow what an object resembles or what a name evokes, rather than the record of its making. They show the missing binding between visual evidence and knowledge described in the introduction.

\section{Conclusion and Future Work}
\Pothen{} tests whether vision-language models can tell who made a cultural object, where, and when from its image alone. Complete attributions are recovered for 2.13\% to 7.87\% of objects, and although images supply evidence and models sometimes name the correct maker, these pieces rarely cohere into one attribution. Reading a cultural object as experts do requires binding what is visible to knowledge of that specific object, which matters more as model descriptions of cultural heritage are reused by other systems and read by people who never see the objects themselves. Future work should extend \Pothen{} to more institutions and to descriptions written by source communities.

\FloatBarrier

\section*{LLM Usage Statement}
OpenAI Codex assisted with evaluation-metric code, recalculation of reported results and diagnostic baselines, manuscript language and layout, result tables, statistical plots, workflow diagrams, and case-summary editing. The authors are responsible for the data, task design, evaluation protocol, analysis, interpretation, and manuscript content.

\section*{Acknowledgements}
Yuechen Jiang is supported by a PhD studentship jointly funded by the Engineering and Physical Sciences Research Council (EPSRC) [grant number EP/Z534912/1] and the Getty Conservation Institute. The University of Manchester provides additional support through a tuition fee waiver.

{\small
\bibliographystyle{ieeenat_fullname}
\bibliography{custom}
}

\clearpage
\appendix
\raggedbottom
\section{Limitations}
\label{appendix:limitations}
Region-stratified sampling aims to broaden cultural coverage and balance representation across the available region-type cells. Region serves as a practical sampling category and is not directly equivalent to culture: each region contains diverse traditions, and cultural relationships often cross geographic boundaries. The 750 objects from two museums reflect their collection coverage, metadata availability, and classification systems. Expanding source institutions, object types, and cultural traditions would further improve coverage. The museum-and-title-only control (Section~\ref{sec:title-ablation}) measures what collection priors and titles alone recover, but it does not separate inference from visual recognition of widely reproduced objects.

Reference granularity and completeness vary across objects. Empty-reference scores assess uncertainty handling, while accuracy on valid outputs should be interpreted alongside the reported invalid-output counts (Appendix~\ref{appendix:metric-definitions}). Judge calibration uses the annotated subset, and broader human validation would strengthen coverage of other outputs. With 50--75 complete records per model drawn from three object subsets, this subset supports aggregate scoring but not a stable model ranking. API model updates affect reproducibility and comparisons.

\section{Ethical Considerations}
\label{appendix:ethics}
\textbf{Museum records and evaluation bias.} Museum metadata reflects collection histories, acquisition practices, and curatorial classification. Records may contain incomplete provenance, uncertain attributions, outdated terminology, or uneven representation of cultural traditions. These factors can shape the reference labels and introduce evaluation bias: a culturally plausible prediction may differ from an institutional description, while agreement with that description may reproduce its assumptions. Scores should therefore be interpreted in relation to the source records and their documented uncertainty. Regional balance supports broader coverage, while the cultural diversity within and across regions requires attention when interpreting regional comparisons.

\textbf{Responsible use of generated metadata.} VLM outputs and LLM-judge decisions require cautious use. Generated titles, dates, origins, and creator attributions can be plausible yet unsupported, and automated scoring can reinforce such errors. Outputs should be treated as candidate interpretations for qualified review, with consultation of source records and relevant cultural expertise before use in cataloguing, provenance research, or public presentation. Uncertainty should remain visible, and generated claims should not silently replace museum records or inform high-stakes valuation, ownership, or restitution decisions. The released results retain a distinction between source references, model predictions, and evaluation decisions.

\textbf{Annotation and prompt design.} The three annotators are authors of this paper. They have no specialist background in the represented traditions and judge compatibility with institution-authored records and descriptions, so their labels measure agreement with the catalogue rather than cultural correctness. All annotators provided informed consent to participate in the annotation and to the use and release of their annotations for this research. In addition to annotating model outputs, they used their annotation experience to design the Judge prompt, including semantic-compatibility and uncertainty-handling rules. This shared involvement informs the evaluation criteria and should be considered when interpreting judge-human agreement. Model names were visible in the annotation workbooks; candidate judges received predictions, reference labels, and museum descriptions, with human decisions and model identity withheld. The annotation protocol and Judge prompt are documented in Appendices~\ref{appendix:annotator-background}--\ref{appendix:annotation-guideline} and Appendix~\ref{appendix:prompt}.

\textbf{Data copyright and attribution.} The Metropolitan Museum of Art releases public-domain artwork images and basic collection data under Creative Commons Zero (CC0). Getty Open Content images marked as public domain are also available under CC0. Source metadata and descriptive text retain the licences specified in the corresponding records; material marked CC BY 4.0 requires attribution. These source licences remain applicable when the data are incorporated into the benchmark. Code and derived annotations follow the licences stated in the release. Reuse should preserve source identifiers, institutional attribution, rights statements, and uncertainty, and distinguish generated metadata from museum-provided information.

\section{Prompt}
\label{appendix:prompt}
The following VLM and judge prompts were used in the experiments.

\textbf{VLM prompt.}
\begin{lstlisting}[basicstyle=\ttfamily\footnotesize,breaklines=true]
You are an expert in cultural heritage and museum collections.

Given one or more images of a cultural heritage object, infer structured metadata that is NOT directly observable and requires cultural, historical, and contextual knowledge.

Predict:
- title
- culture
- period
- origin
- creator or workshop

Important:
- Do NOT describe the image.
- Focus on culturally grounded inference.
- For each uncertain field, explicitly qualify its value with "possibly", "probably", "likely", "may", or "might", provide tentative alternatives, or state "not sure".
- Do not present an uncertain inference as a definite fact. Keep uncertainty expressions inside the corresponding JSON field.

Return STRICT JSON only:
{
  "title": "...",
  "culture": "...",
  "period": "...",
  "origin": "...",
  "creator": "..."
}
\end{lstlisting}

\textbf{Judge prompt.}
The following prompt reproduces \texttt{judge\_prompt.md} in the repository. All four candidates use the same decision rules.
\begin{lstlisting}[basicstyle=\ttfamily\footnotesize,breaklines=true]
You are evaluating metadata generated from an image of a cultural-heritage object.

Compare each predicted field independently with the corresponding reference labels. Return one decision for each of: title, culture, period, origin, and creator.

Decision rules:
- Use 1 when the prediction is fully correct or broadly consistent in meaning and scope with at least one applicable reference label.
- Use 0 when it is incorrect, contradictory, unsupported, has no meaningful overlap, or says unknown/not sure while the reference provides an answer.
- When the corresponding reference-label array is empty, use 1 if the model explicitly answers unknown/not sure/uncertain or gives a substantively uncertain description (for example, "possibly...", "probably...", or an explicit set of alternatives). Use 0 if the model gives a definite specific answer. Judge only the model output's uncertainty in this case; do not infer a missing reference from the sampling region, object type, description, or another field.
- Exact wording is not required. Accept ordinary aliases, spelling or transliteration variants, historical and modern place names, noun/adjectival variants, and equivalent chronological expressions.
- Reference labels are alternatives or complementary acceptable labels. The prediction does not need to reproduce every label.
- Evaluate only the current field. Correct information placed in another field does not make this field correct.
- For period, accept a compatible century, dynasty, reign, date range, or historical period when the temporal range is broadly consistent. Reject a conflicting or materially different period.
- For origin, accept compatible city, region, country, and historical/modern place names when they identify the same origin without contradiction.
- For creator, accept established aliases and the correctly identified person, workshop, factory, manufactory, kiln, studio, or attributed group. Unknown is correct only when the reference explicitly says unknown or anonymous.

The museum description is context only. It may clarify a nonempty reference. When a reference-label array is empty, do not use the description to decide the missing ground truth; apply only the uncertainty rule above.

Give a short reason for each decision. Return only the required JSON object.
\end{lstlisting}
The full-benchmark run additionally requires binary integer scores and per-field reasons in JSON. Format retries reinforce this output contract without changing the decision rules; temperature is 0, the token limit is 1,200, and thinking is disabled.

\textbf{Judge input template.}
The evaluator serialises the following payload. The museum description is context only; model identity and sampling strata are omitted.
\begin{lstlisting}[basicstyle=\ttfamily\footnotesize,breaklines=true]
{
  "museum_description": record["description"],
  "reference_labels": record["reference_labels"],
  "model_output": record["model_output"]
}
\end{lstlisting}

\textbf{Judge output contract.}
The code requires two objects, each containing all five field names. Each score is an integer in \{0,1\}; each reason is a string. The following displays the schema, not an example evaluation.
\begin{lstlisting}[basicstyle=\ttfamily\footnotesize,breaklines=true]
{
  "scores": {
    "title": <0 or 1>,
    "culture": <0 or 1>,
    "period": <0 or 1>,
    "origin": <0 or 1>,
    "creator": <0 or 1>
  },
  "reasons": {
    "title": "<short reason>",
    "culture": "<short reason>",
    "period": "<short reason>",
    "origin": "<short reason>",
    "creator": "<short reason>"
  }
}
\end{lstlisting}
The released evaluation code implements these contracts for cloud and OpenAI-compatible local endpoints, and the release includes the result files of all four candidate judges. Field scores and reasons are retained for audit; no separate holistic label is requested.

\section{Model Specifications}
\label{appendix:model-lists}
\begin{table}[!tbp]
\caption{Candidate judges.}
\label{tab:candidate-judge-models}
\centering
\setlength{\tabcolsep}{3pt}\fontsize{8}{9.5}\selectfont
\begin{tabular}{>{\scriptsize}llp{0.18\columnwidth}r}
\toprule
Model & Abbr. & Org. & Year \\
\midrule
\multicolumn{4}{c}{\textbf{Open-Weight Models}} \\
DeepSeek-V4.1-Flash \cite{deepseek_v4} & DS-J & DeepSeek & 2026 \\
Qwen3-30B-A3B-Instruct-2507 \cite{qwen3_2507} & Qwen-J & Alibaba & 2025 \\
\midrule
\multicolumn{4}{c}{\textbf{Closed-Source Models}} \\
GPT-4.1-mini \cite{openai_gpt41mini} & GPT-J & OpenAI & 2025 \\
Gemini-3.5-Flash-Lite \cite{gemini35flashlite} & Gem-J & Google & 2026 \\
\bottomrule
\end{tabular}

\end{table}

\begin{table}[!tbp]
\caption{Evaluated VLMs.}
\label{tab:vlm-overview}
\centering
\setlength{\tabcolsep}{3pt}\fontsize{8}{9.5}\selectfont
\begin{tabular*}{\columnwidth}{@{\extracolsep{\fill}}>{\scriptsize}llr@{}}
\toprule
Model & Org. & Year \\
\midrule
\multicolumn{3}{c}{\textbf{Open-Weight Models}} \\
Qwen3-VL-8B-Instruct \cite{qwen3_vl} & Alibaba & 2025 \\
Qwen3-VL-32B-Instruct \cite{qwen3_vl} & Alibaba & 2025 \\
Pixtral-12B \cite{pixtral} & Mistral AI & 2024 \\
\midrule
\multicolumn{3}{c}{\textbf{Closed-Source Models}} \\
Qwen-VL-Max \cite{qwen_vl} & Alibaba & 2024 \\
Qwen3-VL-Plus \cite{qwen3_vl} & Alibaba & 2025 \\
Qwen3-VL-Flash \cite{qwen3_vl} & Alibaba & 2025 \\
GPT-4.1-mini \cite{openai_gpt41mini} & OpenAI & 2025 \\
GPT-5.4-mini \cite{openai_gpt54mini} & OpenAI & 2026 \\
Claude Haiku 4.5 \cite{anthropic_claude} & Anthropic & 2025 \\
\bottomrule
\end{tabular*}

\end{table}

\section{Dataset Statistics}
\label{appendix:dataset-statistics}
The evaluation set contains 750 objects in 15 region-type cells. Each available cell contributes 50 objects. Table~\ref{tab:dataset-statistics} gives the complete cell counts. East Asia contains no sculpture cell; consequently, regional and category effects are reported separately rather than interpreted as a fully crossed design. The selection follows the harmonisation rationale in Section~\ref{sec:dataset-construction}: Chinese collection practice gives bronze objects a separate category~\cite{moc2001grading}, but a separate cross-source bronze cell and an East Asian sculpture cell did not satisfy the joint sample and metadata filters.

\begin{table}[H]
\caption{Object counts by region and type.}
\label{tab:dataset-statistics}
\centering
\setlength{\tabcolsep}{3pt}
\fontsize{9}{11}\selectfont
\begin{tabular}{p{0.28\columnwidth}rrrrr}
\toprule
Region & Cer. & Paint. & Metal. & Sculp. & Total \\
\midrule
Americas & 50 & 50 & 50 & 50 & 200 \\
Ancient Mediterranean & 50 & 50 & 50 & 50 & 200 \\
East Asia & 50 & 50 & 50 & 0 & 150 \\
Europe & 50 & 50 & 50 & 50 & 200 \\
\midrule
Total & 200 & 200 & 200 & 150 & 750 \\
\bottomrule
\end{tabular}

\end{table}

The current reference release, \texttt{reference\_labels.json}, contains 750 objects: 328 from the Metropolitan Museum of Art and 422 from Getty. Each entry retains an object identifier, source institution, five reference-label arrays, and a museum description, available for all 422 Getty objects and none of the Met objects. Multiple labels represent acceptable alternatives or complementary expressions. An empty array is distinct from an explicit label such as ``Unknown''. There is no train/development split because the resource is used for evaluation.

Table~\ref{tab:raw-label-space} reports reference availability recomputed from this release. These counts describe the reference data, not model evaluation results. All 126 empty creator references are Met records; every Getty record names a creator, and 200 Getty records list only an unknown, anonymous, or unidentified maker.

\begin{table}[H]
\caption{Reference-label availability across the 750 objects. Empty references follow the uncertainty rule (Appendix~\ref{appendix:annotation-guideline}).}
\label{tab:raw-label-space}
\centering\fontsize{9}{11}\selectfont
\begin{tabular}{lrr}
\toprule
Reference field & Nonempty & Empty \\
\midrule
Title & 750 & 0 \\
Culture & 750 & 0 \\
Period & 672 & 78 \\
Origin & 636 & 114 \\
Creator & 624 & 126 \\
\bottomrule
\end{tabular}

\end{table}

\FloatBarrier
\section{Annotator Background and Annotation System}
\label{appendix:annotator-background}
Three annotators use separate Excel workbooks containing museum descriptions, reference labels, and model outputs. Each workbook contains 75 object rows, each with three models' outputs and adjacent field-level score cells. Thus, a workbook represents 225 model-output assessments. Two copies of a group contain identical objects, references, and predictions; only the assigned annotator differs. Model names are visible in the released workbooks.

\begin{figure}[H]
\centering
\includegraphics[width=\columnwidth,height=0.70\textheight,keepaspectratio]{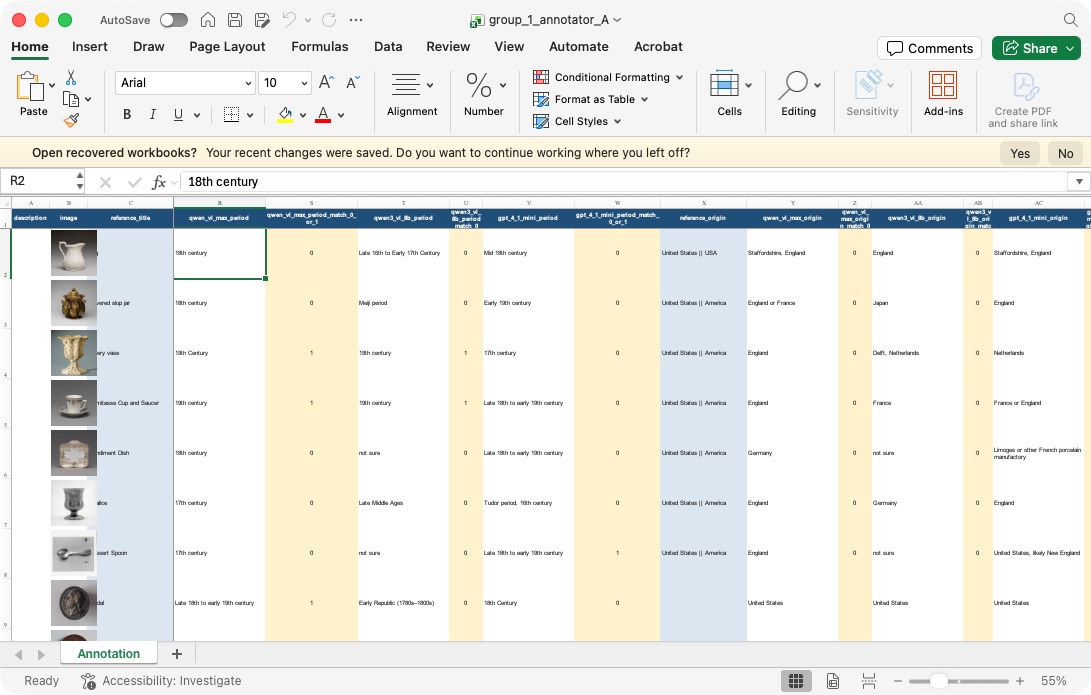}
\caption{Annotation workbook.}
\label{fig:annotation-interface}
\end{figure}

Figure~\ref{fig:annotation-interface} shows the annotation interface.

\begin{table}[H]
\caption{Annotation groups: five objects per region-type cell, three model outputs per object.}
\label{tab:annotation-assignment}
\centering
\setlength{\tabcolsep}{1.5pt}\fontsize{9}{11}\selectfont
\begin{tabular}{c>{\scriptsize}lc}
\toprule
Group & Models & Annotators \\
\midrule
1 & Qwen-VL-Max; Qwen3-VL-8B-Instruct; GPT-4.1-mini & A, B \\
2 & Qwen3-VL-Plus; Qwen3-VL-32B-Instruct; GPT-5.4-mini & B, C \\
3 & Qwen3-VL-Flash; Pixtral-12B; Claude Haiku 4.5 & A, C \\
\bottomrule
\end{tabular}

\end{table}

Each annotator independently annotated two groups (450 outputs). Initial decisions were frozen before disagreement review, and the third annotator independently adjudicated the disputed fields in the remaining group. The final eligible labels and exclusion rules are reported in Section~\ref{sec:human-evaluation} and Appendix~\ref{appendix:metric-definitions}.
\par
\noindent\textbf{Relevant experience.} The annotation team has interdisciplinary backgrounds and experience evaluating language and vision-language models. Annotators used the shared guidelines and independent review protocol described above. Individual educational stages, research specialisations, and experience durations are omitted for anonymous review.

\FloatBarrier
\section{Annotation Guideline}
\label{appendix:annotation-guideline}
The following English guideline follows the bilingual instructions released with the annotation workbooks.
\begin{enumerate}
\item Annotate the five fields independently: \texttt{title}, \texttt{culture}, \texttt{period}, \texttt{origin}, and \texttt{creator}.
\item For each model and field, enter exactly 1 or 0 in the adjacent \texttt{\_match\_0\_or\_1} column. Use 1 when the output is fully or broadly consistent with at least one reference label in meaning and scope. Use 0 when it is incorrect, contradictory, unsupported, has no meaningful overlap, or only says ``unknown/not sure'' when the reference contains an answer.
\item Exact wording is not required. Accept ordinary aliases, spelling or transliteration variants, historical and modern place names, adjectival and noun forms, and equivalent chronological expressions when they refer to the same information.
\item Reference alternatives are separated by \texttt{||}. Agreement with any applicable alternative can receive 1; the output need not reproduce every alternative.
\item Evaluate only the current field. Correct information placed in another field does not make the current field correct.
\item For \texttt{period}, accept a compatible century, dynasty, reign, date range, or historical period when its range is broadly consistent with the reference. Mark a conflicting or materially different period as 0.
\item For \texttt{origin}, a compatible city, region, country, or historical/modern place name may receive 1 when it identifies the same origin without contradiction.
\item For \texttt{creator}, accept established aliases and the correctly identified person, workshop, factory, manufactory, kiln, studio, or attributed group. ``Unknown'' receives 1 only when the reference explicitly indicates an unknown or anonymous creator.
\item If the reference cell is empty, enter 1 when the model output explicitly expresses uncertainty or speculation, such as ``possibly'', ``probably'', ``likely'', ``may'', ``might'', ``unknown'', ``not sure'', or an explicit set of alternatives. Enter 0 for a definite answer without uncertainty. Judge the meaning of the expression, not an incidental keyword. Do not infer a missing reference from the sampling region, object type, description, or another field.
\item Annotate independently. Do not consult the other annotator's decisions before both copies are complete.
\end{enumerate}

\paragraph{Examples.}
Reference culture \texttt{Japanese} with output \texttt{Japan} receives 1. Reference period \texttt{Ming dynasty || 14th--17th century} with output \texttt{Ming period} receives 1. Reference origin \texttt{Strasbourg || Alsace || France} with output \texttt{Alsace, France} receives 1. Reference creator \texttt{Ding kiln} with output \texttt{Unknown} receives 0. A reference title of \texttt{Butter Pat} with an output describing an unrelated vase receives 0.

\paragraph{Adjudication and comparison.}
After both initial assessments are frozen, the third annotator decides disputed fields. A blank human score is missing, not 0. Fields without a valid human label are excluded from judge-human $\kappa$, binary $F_1$ (positive label 1), and agreement. Existing binary labels on empty-reference fields are retained and assess uncertainty handling rather than known-attribute recovery. The original independent labels are retained for human-human reliability analysis (Appendix~\ref{appendix:metric-definitions}).

\section{Details of Metric Definitions}
\label{appendix:metric-definitions}
\subsection{Valid outputs and scoring units}
For one VLM and evaluation slice, let $N$ be the number of attempted outputs ($N=750$ for the full benchmark), $v_i=1$ for a parseable five-field output and 0 otherwise, $\mathcal{I}=\{i:v_i=1\}$, and $V=|\mathcal{I}|$. Invalid outputs ($N-V$) are reported separately and excluded from all accuracy denominators. Structurally valid unknown responses remain valid. A score with $V=0$ is undefined, not zero.

Let $\mathcal{F}$ contain title, culture, period, origin, and creator, with $|\mathcal{F}|=5$. Each $c_{i,f}\in\{0,1\}$ is a field decision: 1 denotes semantic compatibility and 0 incompatibility. For empty references, scores assess uncertainty handling under Appendix~\ref{appendix:prompt}. Write $s_i=\sum_{f\in\mathcal{F}}c_{i,f}$ and use $\mathbf{1}[\cdot]$ for an indicator.

\subsection{Accuracy measures}
\paragraph{Field accuracy.}
The field columns report the proportion of valid outputs with a positive decision:
\begin{equation}
A_f=\frac{1}{V}\sum_{i\in\mathcal{I}}c_{i,f}.
\label{eq:metric-field}
\end{equation}
Mean field accuracy (Mean) is $A_{\mathrm{macro}}=\frac{1}{5}\sum_{f\in\mathcal{F}}A_f$, a macro average of field accuracies rather than class-wise $F_1$.

\paragraph{Reference-aware reporting and independence baselines.}
Let $\mathcal{R}_i=\{k\in\mathcal{F}:m_i^{*k}\ne\varnothing\}$ contain fields with a nonempty reference and $\mathcal{E}_i=\mathcal{F}\setminus\mathcal{R}_i$ contain empty-reference fields.
We report their field-micro scores separately:
\begin{equation}
\begin{aligned}
 \mathrm{Recorded}&=\frac{\sum_{i\in\mathcal{I}}\sum_{k\in\mathcal{R}_i}c_{i,k}}{\sum_{i\in\mathcal{I}}|\mathcal{R}_i|},\\
 \mathrm{Empty}&=\frac{\sum_{i\in\mathcal{I}}\sum_{k\in\mathcal{E}_i}c_{i,k}}{\sum_{i\in\mathcal{I}}|\mathcal{E}_i|}.
\end{aligned}
\end{equation}
Recorded scores agreement with the available museum labels, including explicitly unknown creators. Empty scores uncertainty handling under the empty-reference rule. The latter measures answer behaviour when the reference is absent. The existing Mean, Ctx, and Acc aggregate both components.

For field set $\mathcal{G}\subseteq\mathcal{F}$, the independence baseline is
\begin{equation}
 A_{\mathrm{ind}}(\mathcal{G})=\prod_{k\in\mathcal{G}}\left(\frac{1}{V}\sum_{i\in\mathcal{I}}c_{i,k}\right).
\end{equation}
We also condition on the reference-availability pattern $h_i=(\mathbf{1}[m_i^{*k}\ne\varnothing])_{k\in\mathcal{F}}$:
\begin{equation}
 A_{\mathrm{cond}}(\mathcal{G})=\sum_{h:V_h>0}\frac{V_h}{V}\prod_{k\in\mathcal{G}}\left(\frac{1}{V_h}\sum_{i\in\mathcal{I}:h_i=h}c_{i,k}\right).
\end{equation}
Here $V_h$ counts valid outputs in pattern $h$; patterns with no sampled outputs contribute zero, including in bootstrap resamples. Each baseline is the expected all-correct rate after independently permuting field scores across objects, either globally or within availability patterns, while preserving field marginals. We use $\mathcal{G}=\mathcal{F}$ for Acc and $\mathcal{G}=\mathcal{F}\setminus\{\mathrm{title}\}$ for Ctx. Paired 95\% intervals resample the 750 objects 10,000 times with replacement (NumPy PCG64, seed 20260901); each draw retains all field scores and the model-specific valid-output mask, and recomputes both observed and baseline rates. The null is a descriptive reference for dependence among scores.

\paragraph{Complete-record accuracy (Acc).}
All five fields must receive a positive decision:
\begin{equation}
A_{\mathrm{record}}=\frac{1}{V}\sum_{i\in\mathcal{I}}\mathbf{1}[s_i=5].
\label{eq:metric-record}
\end{equation}
This subset-accuracy criterion~\cite{tsoumakas2007multilabel} is derived from field decisions, not a separate holistic judgement.

\paragraph{Context accuracy (Ctx).}
The four contextual fields $\mathcal{F}_{\mathrm{ctx}}=\mathcal{F}\setminus\{\mathrm{title}\}$ must all receive a positive decision:
\begin{equation}
A_{\mathrm{ctx}}=\frac{1}{V}\sum_{i\in\mathcal{I}}\mathbf{1}\Bigl[\sum_{f\in\mathcal{F}_{\mathrm{ctx}}}c_{i,f}=4\Bigr].
\label{eq:metric-ctx}
\end{equation}
\paragraph{Distribution of compatible fields.}
We report $p_k=V^{-1}\sum_{i\in\mathcal{I}}\mathbf{1}[s_i=k]$ for $k=0,\dots,5$. The distribution sums to one, $p_5=A_{\mathrm{record}}$, and $A_{\mathrm{macro}}=\frac{1}{5}\sum_{k}k\,p_k$, which increases whenever any field decision changes from 0 to 1. Rates are multiplied by 100 for display. Regional metrics use the region-specific valid-output denominator. Empty-reference scores measure uncertainty handling, not known-attribute recovery.

\paragraph{Bootstrap confidence intervals.}
Intervals in Tables~\ref{tab:overall} and~\ref{tab:regional-results} resample objects. Each of $B=10{,}000$ replicates draws 750 objects with replacement (seed 20260901), and every model uses the same draw. With $w_i^{(b)}$ the multiplicity of object $i$ in replicate $b$ and $g_i$ the per-output score ($s_i/5$ for Mean, the Ctx indicator for Ctx, and $\mathbf{1}[s_i=5]$ for Acc), the replicate estimate is $\sum_{i\in\mathcal{I}}w_i^{(b)}g_i\big/\sum_{i\in\mathcal{I}}w_i^{(b)}$. The 95\% interval spans the 2.5th and 97.5th percentiles of the $B$ estimates. Regional intervals restrict both sums to objects in the region.

\paragraph{Pooled human-subset field accuracy.}
For evaluated model $m$, let $\mathcal{E}_m$ contain its eligible object-model--field units with an available human label and binary decisions from all four candidates. If $z_k$ is the human or Judge label, each accuracy column in Table~\ref{tab:human-judge-accuracy} reports
\begin{equation}
A_m^{\mathrm{subset}}(z)=\frac{1}{|\mathcal{E}_m|}\sum_{k\in\mathcal{E}_m}z_k.
\label{eq:metric-subset}
\end{equation}
This positive-field rate is neither complete-record accuracy nor agreement with humans. Valid outputs count distinct outputs contributing at least one eligible field. Missing fields are not counted in the denominator or assigned 0.

\subsection{Paired labels and confusion matrices}
Human--human reliability uses initial paired binary labels before adjudication, pooling A/B, B/C, and A/C as first/second annotators for Groups~1--3. Judge calibration uses the common intersection $\mathcal{E}$ of valid outputs with an available human label and binary decisions from all four candidates. Initial agreements, completed adjudications, and single available initial labels are retained. Missing or nonbinary scores are excluded. Existing binary labels on empty-reference fields are retained; blank cells are missing, not 0. The calibration set has 3,187 units, including 190 empty-reference labels.

Let $x_k$ and $y_k$ be the row and column labels, $n=|\mathcal{E}|$, and
\begin{equation}
C_{ab}=\sum_{k\in\mathcal{E}}\mathbf{1}[x_k=a,\ y_k=b],\quad a,b\in\{0,1\}.
\label{eq:metric-confusion-count}
\end{equation}
With positive class 1, the confusion matrix is
\begin{equation}
\mathbf{C}=\begin{pmatrix}C_{00}&C_{01}\\C_{10}&C_{11}\end{pmatrix}
=\begin{pmatrix}\mathrm{TN}&\mathrm{FP}\\\mathrm{FN}&\mathrm{TP}\end{pmatrix}.
\label{eq:metric-confusion}
\end{equation}
Rows are human labels and columns Judge labels for human--Judge matrices. For human-human matrices, rows and columns are the first and second annotators; neither is ground truth. Labels 0 and 1 denote incompatibility and compatibility. Cells show counts; colour encodes
\begin{equation}
\widetilde{C}_{ab}=100\frac{C_{ab}}{n}.
\label{eq:metric-confusion-colour}
\end{equation}
Normalisation uses all eligible pairs within the panel, not separate rows or columns. Human field panels use field-specific sample counts; all four candidate panels use the same common sample and colour scale.

\subsection{Agreement, binary F1, and Cohen's kappa}
\paragraph{Agreement.}
Observed agreement is the proportion of equal paired labels:
\begin{equation}
p_o=\frac{1}{n}\sum_{k\in\mathcal{E}}\mathbf{1}[x_k=y_k]=\frac{\mathrm{TN}+\mathrm{TP}}{n}.
\label{eq:metric-agreement}
\end{equation}
We report $100p_o$. For human--Judge pairs, this equals binary classification accuracy against the human labels, distinct from the positive-field rate in Equation~\ref{eq:metric-subset}.

\paragraph{Binary F1.}
Positive class 1 denotes compatibility. Precision, recall, and $F_1$ are~\cite{vanrijsbergen1979information}
\begin{align}
P&=\frac{\mathrm{TP}}{\mathrm{TP}+\mathrm{FP}},\\
R&=\frac{\mathrm{TP}}{\mathrm{TP}+\mathrm{FN}},\\
F_1&=\frac{2\mathrm{TP}}{2\mathrm{TP}+\mathrm{FP}+\mathrm{FN}}.
\label{eq:metric-f1}
\end{align}
When defined, $F_1=2PR/(P+R)$. We report $100F_1$, computed from pooled counts, not class-macro or support-weighted $F_1$. The count formula applies even if precision or recall has a zero denominator. If the $F_1$ denominator is zero for a nonempty comparison, $F_1$ is assigned 0. Empty comparisons have no reported statistic.

\paragraph{Cohen's kappa.}
Let $r_a=\sum_bC_{ab}/n$ and $q_a=\sum_bC_{ba}/n$ be the row and column marginal proportions. Chance-expected agreement and Cohen's $\kappa$ are~\cite{cohen1960coefficient}
\begin{align}
p_e&=r_0q_0+r_1q_1,\label{eq:metric-chance}\\
\kappa&=\frac{p_o-p_e}{1-p_e}.\label{eq:metric-kappa}
\end{align}
For $p_e\ne1$, $\kappa=1$ denotes perfect agreement, $\kappa=0$ agreement at the marginal chance level, and negative values agreement below that level. If $p_e=1$, $\kappa$ is undefined. Judge selection uses the highest pooled $\kappa$ before rounding.

\paragraph{Overall and field-wise aggregation.}
Overall agreement, $F_1$, and $\kappa$ are recomputed after pooling eligible units across fields, objects, and models; they are not averages of field-wise statistics. Each field uses its own eligible subset and the same formulas. All candidates share identical pairing and missing-value rules. Agreement and $F_1$ are percentages; $\kappa$ is unitless.

\paragraph{Worked example.}
The candidate-stage DS-J run has $\mathrm{TN}=1{,}596$, $\mathrm{FP}=105$, $\mathrm{FN}=231$, and $\mathrm{TP}=1{,}255$, with $n=3{,}187$. Equations~\ref{eq:metric-agreement}, \ref{eq:metric-f1}, and~\ref{eq:metric-kappa} give 89.46\% agreement, 88.19\% $F_1$, and $\kappa=0.787$, matching Table~\ref{tab:judge-selection}.

\FloatBarrier
\begingroup
\captionsetup{skip=3pt}
\makeatletter
\def\cvprsubsection{\@startsection{subsection}{2}{\z@}{-6pt}{4pt}{\elvbf}}
\renewcommand\paragraph{\@startsection{paragraph}{4}{\z@}{6pt}{-1em}{\normalfont\normalsize\bfseries}}
\makeatother
\section{Additional Results and Model Details}
\label{appendix:additional-results}
\subsection{Candidate and Full-Run DS-J Comparison}
\label{sec:dsj-run-stability}
Both stored judge runs are paired with the 3,187 eligible fields in the released final human annotations, using identical object-model--field keys.

On the final set, the runs agree on 3,042 fields (95.45\%; inter-run $\kappa=0.907$, 95\% CI [0.891, 0.923]). The 145 changes (4.55\%) comprise 73 from 0 to 1 and 72 from 1 to 0; 50 correct and 95 introduce a disagreement with humans. Full-run human agreement is 88.05\%, binary $F_1$ 86.62\%, and $\kappa=0.759$ [0.729, 0.786], compared with candidate $\kappa=0.787$ [0.760, 0.813] (Table~\ref{tab:human-judge-record}; Figure~\ref{fig:dsj-run-comparison}). The stages differ in format enforcement and retries. Selection and the worked example use the candidate run; matched-record and full-benchmark results use the full run.
\begin{figure}[H]
\centering
\includegraphics[width=\columnwidth,height=0.70\textheight,keepaspectratio]{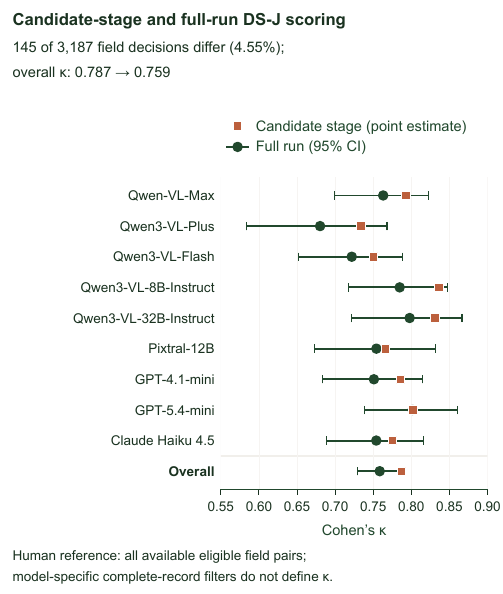}
\caption{Candidate-stage and full-run DS-J agreement with the final human reference. Squares: candidate-stage $\kappa$; circles and whiskers: full-run $\kappa$ and 95\% object-cluster bootstrap CIs. The runs differ on 145 of 3,187 field decisions (4.55\%).}
\label{fig:dsj-run-comparison}
\end{figure}

\subsection{Per-Model Human and Judge Agreement}
Of 675 reviewed outputs, 19 are invalid and 584 have all five human labels. Table~\ref{tab:human-judge-record} reports $\kappa$ on all 3,187 available field pairs; Figure~\ref{fig:human-dsj-record} separates this population from Acc on complete records (Table~\ref{tab:matched-subset-metrics}). Missing labels are excluded without imputing zero. DS-J has a lower compatible-field proportion than humans (42.70\% versus 46.63\%); label agreement is a separate quantity.

\paragraph{Kappa confidence intervals.}
We use 20,000 object-cluster bootstrap resamples within the three annotation groups (75 objects each), retaining all eligible fields and model outputs of each sampled object. All comparisons share PCG64 draws with seed 20261109 and sorted object IDs. Kappa is recomputed from pooled confusion counts; 95\% intervals use the 2.5th and 97.5th percentiles with linear interpolation. All reported replicates have defined kappa.
\begin{table}[H]
\caption{Human and full-run DS-J field agreement on all 3,187 available pairs. $n_f$: paired fields used for $\kappa$; candidate $\kappa$: judge-selection run. CIs use 20,000 stratified object-cluster bootstrap resamples. Missing labels are excluded. Record Acc is reported in Table~\ref{tab:matched-subset-metrics}.}
\label{tab:human-judge-record}
\centering
\fontsize{8}{9}\selectfont
\setlength{\tabcolsep}{2pt}
\textbf{(a) Field-level agreement}\\[2pt]
\begin{tabular*}{\columnwidth}{@{\extracolsep{\fill}}>{\scriptsize}lrrc@{}}
\toprule
Model & $n_f$ & Candidate $\kappa$ & Full-run $\kappa$ [95\% CI] \\
\midrule
Qwen-VL-Max & 339 & 0.793 & 0.763 {\scriptsize [0.699, 0.823]} \\
Qwen3-VL-Plus & 345 & 0.734 & 0.680 {\scriptsize [0.583, 0.768]} \\
Qwen3-VL-Flash & 360 & 0.750 & 0.722 {\scriptsize [0.652, 0.788]} \\
Qwen3-VL-8B-Instruct & 344 & 0.836 & 0.785 {\scriptsize [0.717, 0.847]} \\
Qwen3-VL-32B-Instruct & 375 & 0.831 & 0.798 {\scriptsize [0.721, 0.867]} \\
Pixtral-12B & 335 & 0.766 & 0.754 {\scriptsize [0.673, 0.831]} \\
GPT-4.1-mini & 344 & 0.786 & 0.751 {\scriptsize [0.684, 0.815]} \\
GPT-5.4-mini & 375 & 0.802 & 0.802 {\scriptsize [0.738, 0.861]} \\
Claude Haiku 4.5 & 370 & 0.775 & 0.754 {\scriptsize [0.688, 0.816]} \\
\midrule
Overall & 3,187 & 0.787 & 0.759 {\scriptsize [0.729, 0.786]} \\
\bottomrule
\end{tabular*}
\par\smallskip
\textbf{(b) All available field pairs ($n_f=3{,}187$)}\\[2pt]
\begin{tabular*}{\columnwidth}{@{\extracolsep{\fill}}lrr@{}}
\toprule
Statistic & Human & Full-run DS-J \\
\midrule
Field accuracy (\%) & 46.63 & 42.70 \\
Human--DS-J agreement (\%) & \multicolumn{2}{c}{88.05} \\
Human--DS-J binary $F_1$ (\%) & \multicolumn{2}{c}{86.62} \\
\bottomrule
\end{tabular*}
\end{table}

\begin{figure}[!t]
\centering
\includegraphics[width=\columnwidth,height=0.70\textheight,keepaspectratio]{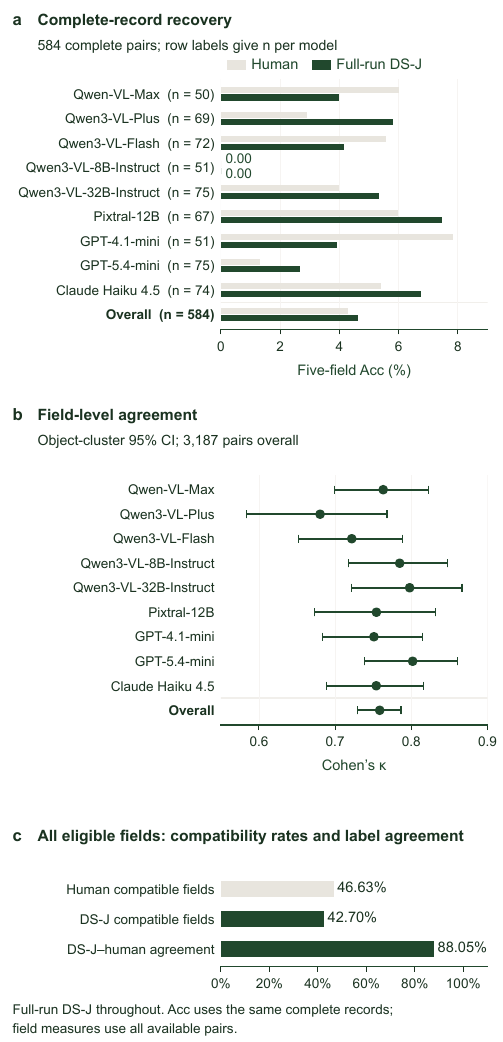}
\caption{Human and full-run DS-J comparison. (a) Five-field Acc on the same 584 complete records, with model-specific sample sizes. (b) Field-level $\kappa$ and 95\% object-cluster bootstrap CIs on all available pairs (3,187 pooled). (c) Compatible-field proportions and binary label agreement on those pairs.}
\label{fig:human-dsj-record}
\end{figure}

\subsection{Bootstrap Ranking Stability}
Table~\ref{tab:ranking-stability} and Figure~\ref{fig:ranking-stability} report 20,000 object-cluster bootstrap resamples stratified by annotation group, retaining each sampled object's outputs and fields. The pooled full-run DS-J$-$human Acc difference spans zero ($+0.34$ points [$-1.39$, 2.13]), whereas model Acc rankings remain unstable (Spearman $\rho=0.283$ [0.000, 0.912]). Each model has 50--75 complete records and few all-five-field successes across three object subsets, so cross-group rankings also reflect sample composition. Acc first-place frequencies include ties; field-accuracy frequencies require sole first place.
\begin{table}[H]
\caption{Ranking stability from 20,000 object-cluster bootstrap replicates stratified by annotation group, retaining all outputs and fields of sampled objects. Brackets: 95\% CIs; pp: percentage points. Acc first-place frequencies include ties; field-accuracy frequencies require sole first place.}
\label{tab:ranking-stability}
\centering
\fontsize{8}{9}\selectfont
\setlength{\tabcolsep}{2pt}
\begin{tabular*}{\columnwidth}{@{\extracolsep{\fill}}>{\raggedright\arraybackslash}p{.54\columnwidth}>{\raggedright\arraybackslash}p{.42\columnwidth}@{}}
\toprule
Statistic & Result \\
\midrule
\multicolumn{2}{l}{\textbf{Record Acc: 584 complete paired records}} \\
Human point-estimate leader & \mbox{\scriptsize GPT-4.1-mini}: 7.84\% \\
DS-J point-estimate leader & \mbox{\scriptsize Pixtral-12B}: 7.46\% \\
Human leader ranks first, including ties & 38.73\% \\
DS-J leader ranks first, including ties & 34.37\% \\
Human--DS-J rank Spearman $\rho$ [95\% CI] & 0.283 [0.000, 0.912] \\
Human Acc: \mbox{\scriptsize GPT-4.1-mini} $-$ \mbox{\scriptsize Pixtral-12B} & $+1.87$ pp [$-7.23$, $11.62$] \\
Pooled Acc: full-run DS-J $-$ human & $+0.34$ pp [$-1.39$, $2.13$] \\
\midrule
\multicolumn{2}{l}{\textbf{Reported field-accuracy ranking}} \\
Point-estimate leader under both scorings & \mbox{\scriptsize Qwen3-VL-Flash} \\
Human leader ranks sole first & 51.38\% \\
DS-J leader ranks sole first & 82.77\% \\
Human--DS-J rank Spearman $\rho$ & 0.650 \\
\bottomrule
\end{tabular*}
\end{table}

\begin{figure}[!t]
\centering
\includegraphics[width=0.97\columnwidth,height=0.65\textheight,keepaspectratio]{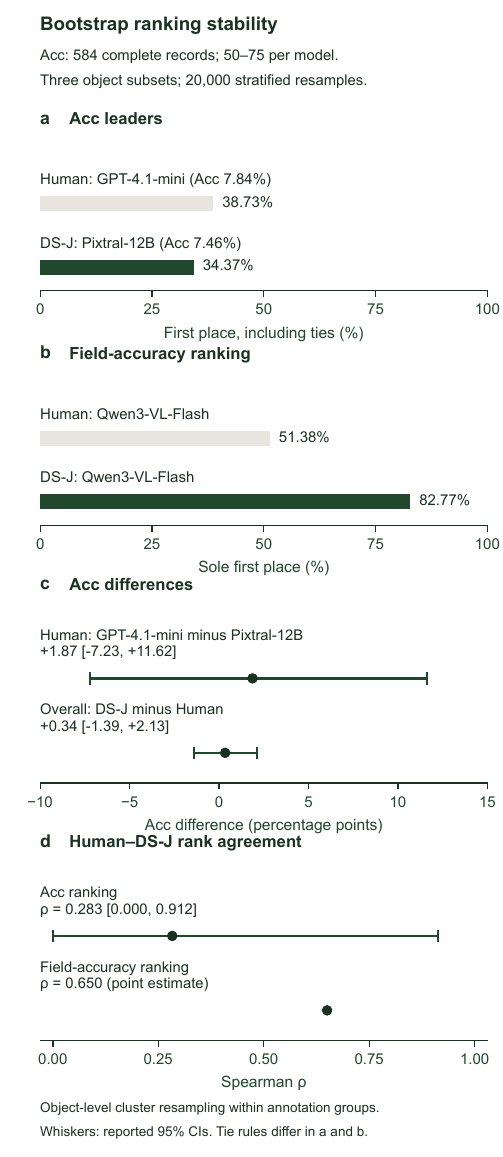}
\caption{Ranking stability from 20,000 object-cluster bootstrap resamples stratified by annotation group. (a) Acc leaders\textquotesingle{} first-place frequencies, including ties, on 584 complete records. (b) Sole-first frequencies for the reported field-accuracy ranking. (c) Acc differences in percentage points. (d) Human--DS-J rank correlations. Whiskers: reported 95\% CIs; field-accuracy $\rho$ is a point estimate. DS-J uses the full run.}
\label{fig:ranking-stability}
\end{figure}

\subsection{Full-Benchmark Distribution and Metric Comparison}
Table~\ref{tab:compatible-field-distribution} provides the exact counts plotted in Figure~\ref{fig:compatible-field-distribution}. These 6,542 valid outputs are distinct from the 584 complete human-labelled records in Table~\ref{tab:matched-subset-metrics}. Figure~\ref{fig:field-record-results} visualises the Mean, Ctx, and Acc results in Table~\ref{tab:overall}.

\begin{table}[!tp]
\caption{Regional DS-J results; notation as in Table~\ref{tab:overall}.}
\label{tab:regional-results}
\renewcommand{\arraystretch}{0.94}
\centering
\fontsize{8}{9.5}\selectfont
\setlength{\tabcolsep}{3pt}
\begin{minipage}{\linewidth}
\centering
\textbf{(a) Americas}\\[0.25em]
\begin{tabular*}{\textwidth}{@{\extracolsep{\fill}}>{\scriptsize}lrrrrrrrrr@{}}
\toprule
Model & Inv. & Mean & Ctx & Acc & Title & Cul. & Per. & Ori. & Cre. \\
\midrule
Qwen-VL-Max & 6 & 36.91\,{\scriptsize[32.89, 40.99]} & 10.31\,{\scriptsize[6.25, 14.85]} & 6.19\,{\scriptsize[3.02, 9.79]} & 31.96 & 39.69 & 56.70 & 35.05 & 21.13 \\
Qwen3-VL-Plus & 6 & 36.70\,{\scriptsize[32.50, 41.09]} & 8.76\,{\scriptsize[5.00, 12.87]} & 6.19\,{\scriptsize[3.02, 9.78]} & 33.51 & 41.24 & 47.42 & 39.18 & 22.16 \\
Qwen3-VL-Flash & 6 & 44.54\,{\scriptsize[40.69, 48.52]} & 11.34\,{\scriptsize[7.14, 15.98]} & 6.70\,{\scriptsize[3.40, 10.38]} & 39.69 & 51.03 & 61.34 & 46.39 & 24.23 \\
Qwen3-VL-8B-Instruct & 0 & 33.10\,{\scriptsize[29.25, 37.04]} & 6.50\,{\scriptsize[3.30, 10.15]} & 3.00\,{\scriptsize[0.97, 5.59]} & 31.50 & 31.00 & 41.00 & 39.50 & 22.50 \\
Qwen3-VL-32B-Instruct & 0 & 36.80\,{\scriptsize[32.55, 41.17]} & 11.00\,{\scriptsize[6.90, 15.61]} & 7.00\,{\scriptsize[3.69, 10.78]} & 26.50 & 42.50 & 53.50 & 38.00 & 23.50 \\
Pixtral-12B & 4 & 27.86\,{\scriptsize[24.37, 31.54]} & 3.57\,{\scriptsize[1.10, 6.40]} & 2.04\,{\scriptsize[0.48, 4.19]} & 33.16 & 22.96 & 41.33 & 28.57 & 13.27 \\
GPT-4.1-mini & 0 & 40.60\,{\scriptsize[36.63, 44.77]} & 11.00\,{\scriptsize[6.83, 15.68]} & 6.00\,{\scriptsize[2.96, 9.52]} & 34.50 & 43.00 & 65.00 & 39.50 & 21.00 \\
GPT-5.4-mini & 0 & 36.70\,{\scriptsize[32.80, 40.74]} & 8.50\,{\scriptsize[4.85, 12.63]} & 4.50\,{\scriptsize[1.90, 7.65]} & 36.50 & 38.50 & 53.50 & 37.00 & 18.00 \\
Claude Haiku 4.5 & 2 & 35.56\,{\scriptsize[31.96, 39.37]} & 4.55\,{\scriptsize[1.95, 7.61]} & 2.53\,{\scriptsize[0.52, 4.95]} & 33.33 & 33.33 & 60.61 & 35.86 & 14.65 \\
\bottomrule
\end{tabular*}
\par\smallskip
\textbf{(b) Ancient Mediterranean}\\[0.25em]
\begin{tabular*}{\textwidth}{@{\extracolsep{\fill}}>{\scriptsize}lrrrrrrrrr@{}}
\toprule
Model & Inv. & Mean & Ctx & Acc & Title & Cul. & Per. & Ori. & Cre. \\
\midrule
Qwen-VL-Max & 13 & 47.06\,{\scriptsize[43.28, 50.73]} & 6.95\,{\scriptsize[3.45, 10.73]} & 2.67\,{\scriptsize[0.55, 5.16]} & 33.69 & 64.71 & 47.06 & 43.85 & 45.99 \\
Qwen3-VL-Plus & 13 & 36.58\,{\scriptsize[32.27, 40.90]} & 9.09\,{\scriptsize[5.21, 13.37]} & 3.74\,{\scriptsize[1.18, 6.59]} & 24.60 & 63.64 & 36.90 & 33.69 & 24.06 \\
Qwen3-VL-Flash & 13 & 46.42\,{\scriptsize[42.13, 50.57]} & 15.51\,{\scriptsize[10.58, 20.86]} & 5.88\,{\scriptsize[2.73, 9.44]} & 27.27 & 66.31 & 43.85 & 40.64 & 54.01 \\
Qwen3-VL-8B-Instruct & 0 & 35.30\,{\scriptsize[31.85, 38.74]} & 4.50\,{\scriptsize[1.88, 7.54]} & 1.50\,{\scriptsize[0.00, 3.32]} & 18.50 & 48.00 & 36.50 & 30.50 & 43.00 \\
Qwen3-VL-32B-Instruct & 0 & 40.20\,{\scriptsize[36.42, 44.04]} & 7.00\,{\scriptsize[3.62, 10.78]} & 2.00\,{\scriptsize[0.47, 4.15]} & 24.00 & 53.00 & 37.50 & 43.00 & 43.50 \\
Pixtral-12B & 22 & 47.64\,{\scriptsize[43.31, 52.04]} & 18.54\,{\scriptsize[12.88, 24.47]} & 7.30\,{\scriptsize[3.68, 11.47]} & 25.28 & 57.87 & 38.20 & 43.82 & 73.03 \\
GPT-4.1-mini & 0 & 45.60\,{\scriptsize[41.49, 49.68]} & 11.50\,{\scriptsize[7.30, 16.11]} & 4.00\,{\scriptsize[1.48, 6.90]} & 31.00 & 63.50 & 45.00 & 41.50 & 47.00 \\
GPT-5.4-mini & 0 & 50.90\,{\scriptsize[46.77, 55.09]} & 19.50\,{\scriptsize[14.21, 25.26]} & 9.00\,{\scriptsize[5.24, 13.07]} & 33.00 & 68.50 & 45.50 & 47.50 & 60.00 \\
Claude Haiku 4.5 & 7 & 45.29\,{\scriptsize[41.08, 49.46]} & 17.62\,{\scriptsize[12.44, 23.12]} & 9.84\,{\scriptsize[5.91, 14.15]} & 25.39 & 55.96 & 34.72 & 38.86 & 71.50 \\
\bottomrule
\end{tabular*}
\par\smallskip
\textbf{(c) East Asia}\\[0.25em]
\begin{tabular*}{\textwidth}{@{\extracolsep{\fill}}>{\scriptsize}lrrrrrrrrr@{}}
\toprule
Model & Inv. & Mean & Ctx & Acc & Title & Cul. & Per. & Ori. & Cre. \\
\midrule
Qwen-VL-Max & 0 & 60.13\,{\scriptsize[55.95, 64.30]} & 23.33\,{\scriptsize[16.67, 30.25]} & 11.33\,{\scriptsize[6.47, 16.78]} & 37.33 & 75.33 & 60.67 & 73.33 & 54.00 \\
Qwen3-VL-Plus & 0 & 59.33\,{\scriptsize[55.00, 63.56]} & 19.33\,{\scriptsize[13.25, 25.83]} & 10.67\,{\scriptsize[5.97, 15.87]} & 40.00 & 84.00 & 48.00 & 76.67 & 48.00 \\
Qwen3-VL-Flash & 0 & 65.20\,{\scriptsize[61.24, 69.02]} & 38.00\,{\scriptsize[30.28, 45.95]} & 14.67\,{\scriptsize[9.21, 20.55]} & 32.67 & 80.00 & 70.00 & 79.33 & 64.00 \\
Qwen3-VL-8B-Instruct & 0 & 45.47\,{\scriptsize[40.90, 49.93]} & 13.33\,{\scriptsize[8.16, 19.01]} & 3.33\,{\scriptsize[0.69, 6.57]} & 19.33 & 54.67 & 44.67 & 56.67 & 52.00 \\
Qwen3-VL-32B-Instruct & 0 & 51.87\,{\scriptsize[47.22, 56.53]} & 21.33\,{\scriptsize[15.15, 28.00]} & 8.67\,{\scriptsize[4.49, 13.43]} & 29.33 & 62.67 & 49.33 & 61.33 & 56.67 \\
Pixtral-12B & 36 & 60.35\,{\scriptsize[55.08, 65.63]} & 35.96\,{\scriptsize[27.18, 44.90]} & 16.67\,{\scriptsize[10.00, 23.81]} & 25.44 & 68.42 & 65.79 & 72.81 & 69.30 \\
GPT-4.1-mini & 0 & 62.67\,{\scriptsize[59.03, 66.30]} & 25.33\,{\scriptsize[18.79, 32.62]} & 10.00\,{\scriptsize[5.52, 15.03]} & 40.00 & 80.00 & 62.67 & 75.33 & 55.33 \\
GPT-5.4-mini & 0 & 62.40\,{\scriptsize[58.01, 66.75]} & 31.33\,{\scriptsize[24.00, 38.82]} & 14.67\,{\scriptsize[9.33, 20.67]} & 42.67 & 76.67 & 63.33 & 72.67 & 56.67 \\
Claude Haiku 4.5 & 4 & 58.49\,{\scriptsize[53.33, 63.68]} & 32.19\,{\scriptsize[24.65, 39.86]} & 18.49\,{\scriptsize[12.50, 25.00]} & 40.41 & 69.18 & 53.42 & 65.07 & 64.38 \\
\bottomrule
\end{tabular*}
\par\smallskip
\textbf{(d) Europe}\\[0.25em]
\begin{tabular*}{\textwidth}{@{\extracolsep{\fill}}>{\scriptsize}lrrrrrrrrr@{}}
\toprule
Model & Inv. & Mean & Ctx & Acc & Title & Cul. & Per. & Ori. & Cre. \\
\midrule
Qwen-VL-Max & 19 & 45.19\,{\scriptsize[41.16, 49.27]} & 8.29\,{\scriptsize[4.46, 12.57]} & 4.42\,{\scriptsize[1.67, 7.61]} & 38.12 & 55.25 & 60.22 & 51.38 & 20.99 \\
Qwen3-VL-Plus & 20 & 40.22\,{\scriptsize[35.86, 44.55]} & 8.33\,{\scriptsize[4.57, 12.65]} & 2.78\,{\scriptsize[0.58, 5.42]} & 31.67 & 65.56 & 38.89 & 48.33 & 16.67 \\
Qwen3-VL-Flash & 20 & 46.22\,{\scriptsize[42.24, 50.12]} & 7.22\,{\scriptsize[3.63, 11.17]} & 2.22\,{\scriptsize[0.52, 4.58]} & 30.00 & 66.67 & 51.11 & 60.56 & 22.78 \\
Qwen3-VL-8B-Instruct & 0 & 33.10\,{\scriptsize[29.21, 36.92]} & 6.00\,{\scriptsize[2.94, 9.52]} & 1.00\,{\scriptsize[0.00, 2.60]} & 18.50 & 44.00 & 37.50 & 45.00 & 20.50 \\
Qwen3-VL-32B-Instruct & 0 & 34.90\,{\scriptsize[30.69, 39.22]} & 8.00\,{\scriptsize[4.52, 11.93]} & 2.50\,{\scriptsize[0.52, 4.89]} & 24.00 & 46.50 & 41.50 & 41.50 & 21.00 \\
Pixtral-12B & 12 & 38.51\,{\scriptsize[34.43, 42.60]} & 7.45\,{\scriptsize[3.85, 11.43]} & 1.60\,{\scriptsize[0.00, 3.63]} & 28.19 & 47.87 & 44.68 & 49.47 & 22.34 \\
GPT-4.1-mini & 0 & 49.10\,{\scriptsize[44.86, 53.40]} & 16.50\,{\scriptsize[11.48, 21.84]} & 11.00\,{\scriptsize[6.81, 15.56]} & 42.00 & 59.50 & 64.00 & 53.00 & 27.00 \\
GPT-5.4-mini & 0 & 41.40\,{\scriptsize[37.22, 45.50]} & 10.50\,{\scriptsize[6.40, 14.90]} & 5.00\,{\scriptsize[2.11, 8.19]} & 29.50 & 57.00 & 42.00 & 52.50 & 26.00 \\
Claude Haiku 4.5 & 5 & 40.82\,{\scriptsize[36.84, 44.80]} & 12.82\,{\scriptsize[8.29, 17.73]} & 1.54\,{\scriptsize[0.00, 3.43]} & 22.56 & 57.44 & 49.74 & 49.23 & 25.13 \\
\bottomrule
\end{tabular*}
\par\smallskip
\end{minipage}
\end{table}

\begin{table}[H]
\caption{Compatible-field counts $s_i$ for 6,542 valid full-run DS-J outputs; 208 invalid outputs are excluded. Mean field accuracy is 44.01\%. Shares are rounded.}
\label{tab:compatible-field-distribution}
\centering
\fontsize{9}{11}\selectfont
\setlength{\tabcolsep}{1.5pt}
\begin{tabular*}{\columnwidth}{@{\extracolsep{\fill}}lrrrrrr@{}}
\toprule
$s_i$ & 0 & 1 & 2 & 3 & 4 & 5 \\
\midrule
Outputs & 1,002 & 1,443 & 1,225 & 1,378 & 1,101 & 393 \\
Share (\%) & 15.32 & 22.06 & 18.73 & 21.06 & 16.83 & 6.01 \\
\bottomrule
\end{tabular*}
\end{table}

\begin{table}[H]
\caption{Pooled field accuracy (\%) on shared human-judge pairs, using candidate-stage judge scores.}
\label{tab:human-judge-accuracy}
\centering
\setlength{\tabcolsep}{.7pt}\fontsize{9}{11}\selectfont
\begin{tabular*}{\columnwidth}{@{\extracolsep{\fill}}>{\scriptsize}lrrrrrr@{}}
\toprule
Model & Valid & Human & GPT-J & DS-J & Gem-J & Qwen-J \\
\midrule
Qwen-VL-Max & 74 & 48.97 & 37.46 & 42.18 & 38.64 & 38.94 \\
Qwen3-VL-Plus & 69 & 44.93 & 38.84 & 40.00 & 42.03 & 32.17 \\
Qwen3-VL-Flash & 72 & 52.50 & 48.89 & 50.00 & 53.61 & 47.50 \\
Qwen3-VL-8B-Instruct & 75 & 38.08 & 30.81 & 34.01 & 32.27 & 36.34 \\
Qwen3-VL-32B-Instruct & 75 & 44.00 & 43.73 & 41.60 & 41.33 & 48.00 \\
Pixtral-12B & 67 & 47.16 & 41.79 & 45.07 & 45.67 & 38.21 \\
GPT-4.1-mini & 75 & 51.45 & 41.86 & 42.44 & 43.02 & 39.83 \\
GPT-5.4-mini & 75 & 47.73 & 44.80 & 44.80 & 44.80 & 41.33 \\
Claude Haiku 4.5 & 74 & 44.86 & 36.76 & 43.51 & 42.97 & 37.03 \\
\bottomrule
\end{tabular*}

\end{table}

The matched human-scored subset is summarised in Section~\ref{sec:human-evaluation}; the analyses above provide run-specific agreement and ranking details.

\subsection{Reference-Aware Scores and Independent-Field Diagnostics}

\label{appendix:reference-diagnostics}
Three objects with a named creator also allow unknown alternatives; both text-only models receive three positive decisions for unknown answers on these objects. Scores for the named-reference group therefore measure compatibility with its reference alternatives.
In the museum-and-title condition, answered cases have culture accuracy 63.4\% and 75.3\% (GPT-4.1-mini and GPT-5.4-mini), compared with 69.6\% and 66.0\% under image input on the same objects. For abstained cases, image-condition culture accuracy is 51.1\% and 52.4\%.

Table~\ref{tab:field-reference-split} reports the per-field reference split. Table~\ref{tab:context-independence} gives the four-field counterpart of the Acc baseline, and Table~\ref{tab:creator-ablation-split} separates creator reference groups in the input ablation.

\begin{table}[!t]

\caption{Per-field full-run DS-J scores by reference availability. Entries are percentage (eligible count). R: recorded, including explicit unknown labels; E: empty-reference uncertainty handling. Title and culture have no empty references. Invalid outputs are excluded.}

\label{tab:field-reference-split}

\centering\fontsize{8}{10}\selectfont

\setlength{\tabcolsep}{2pt}

\begin{tabular*}{\textwidth}{@{\extracolsep{\fill}}lrrrrrrrr@{}}

\toprule

Model & Title R & Culture R & Period R & Period E & Origin R & Origin E & Creator R & Creator E \\

\midrule

Qwen-VL-Max & 35.11 (712) & 57.72 (712) & 55.42 (637) & 60.00 (75) & 50.58 (601) & 44.14 (111) & 23.34 (587) & 87.20 (125) \\

Qwen3-VL-Plus & 32.07 (711) & 62.31 (711) & 43.08 (636) & 38.67 (75) & 49.00 (600) & 42.34 (111) & 15.87 (586) & 77.60 (125) \\

Qwen3-VL-Flash & 32.49 (711) & 65.12 (711) & 55.35 (636) & 61.33 (75) & 55.33 (600) & 55.86 (111) & 28.67 (586) & 93.60 (125) \\

Qwen3-VL-8B & 22.13 (750) & 43.73 (750) & 37.95 (672) & 53.85 (78) & 40.41 (636) & 50.88 (114) & 23.08 (624) & 84.13 (126) \\

Qwen3-VL-32B & 25.73 (750) & 50.40 (750) & 43.75 (672) & 57.69 (78) & 43.40 (636) & 53.51 (114) & 23.24 (624) & 92.06 (126) \\

Pixtral-12B & 28.40 (676) & 46.75 (676) & 45.29 (605) & 47.89 (71) & 44.64 (569) & 52.34 (107) & 32.12 (576) & 92.00 (100) \\

GPT-4.1-mini & 36.67 (750) & 60.27 (750) & 59.52 (672) & 53.85 (78) & 52.36 (636) & 42.11 (114) & 26.12 (624) & 87.30 (126) \\

GPT-5.4-mini & 34.93 (750) & 59.07 (750) & 50.30 (672) & 50.00 (78) & 52.67 (636) & 42.11 (114) & 29.65 (624) & 85.71 (126) \\

Claude Haiku 4.5 & 29.78 (732) & 52.87 (732) & 49.09 (658) & 52.70 (74) & 45.25 (621) & 50.45 (111) & 32.02 (609) & 93.50 (123) \\

\bottomrule

\end{tabular*}

\end{table}

\begin{table}[t]
\centering
\caption{Observed four-context-field Ctx and independence baselines (\%). Notation and resampling match Table~\ref{tab:independence}.}
\label{tab:context-independence}
\fontsize{8}{10}\selectfont\setlength{\tabcolsep}{2pt}
\begin{tabular}{lrrrr}
\toprule
Model & Ctx & Ind. & Cond. & $\Delta$ [95\% CI] \\
\midrule
Qwen-VL-Max & 11.66 & 5.53 & 6.79 & +4.87 [+3.13, +6.53] \\
Qwen3-VL-Plus & 10.97 & 3.40 & 4.77 & +6.20 [+4.35, +7.95] \\
Qwen3-VL-Flash & 17.02 & 8.10 & 9.33 & +7.69 [+5.81, +9.41] \\
Qwen3-VL-8B & 7.20 & 2.42 & 2.57 & +4.63 [+3.11, +6.16] \\
Qwen3-VL-32B & 11.20 & 3.56 & 3.86 & +7.34 [+5.57, +8.97] \\
Pixtral-12B & 14.05 & 4.00 & 5.91 & +8.14 [+6.20, +9.89] \\
GPT-4.1-mini & 15.47 & 6.57 & 7.34 & +8.13 [+6.19, +9.95] \\
GPT-5.4-mini & 16.53 & 5.92 & 6.53 & +10.01 [+8.02, +11.78] \\
Claude Haiku 4.5 & 15.71 & 5.10 & 5.95 & +9.76 [+7.83, +11.54] \\
\bottomrule
\end{tabular}
\end{table}

\begin{table}[t]
\centering
\caption{Creator ablation decomposed by reference status. Accuracies are percentages; Named includes at least one named attribution (423 objects), possibly with unknown alternatives. Unknown contains only unknown-maker labels (201); Empty contains no creator label (126).}
\label{tab:creator-ablation-split}
\fontsize{8}{10}\selectfont\setlength{\tabcolsep}{2pt}
\begin{tabular}{llrrrr}
\toprule
Model & Input & All & Named & Unknown & Empty \\
\midrule
GPT-4.1-mini & Images & 36.40 & 15.60 & 48.26 & 87.30 \\
GPT-4.1-mini & Museum/title & 43.73 & 1.18 & 98.01 & 100.00 \\
GPT-5.4-mini & Images & 39.07 & 13.48 & 63.68 & 85.71 \\
GPT-5.4-mini & Museum/title & 44.13 & 2.36 & 97.01 & 100.00 \\
\bottomrule
\end{tabular}
\end{table}

\subsection{Judge and VLM Specifications}
\label{appendix:model-specifications}

\begin{table}[H]
\caption{Evaluated endpoints (n/r: parameter count not reported).}
\label{tab:models}
\centering
\fontsize{9}{11}\selectfont
\begin{tabular}{>{\scriptsize}lll}
\toprule
Model & Access & Scale \\
\midrule
Qwen-VL-Max~\cite{qwen_vl} & API & n/r \\
Qwen3-VL-Plus~\cite{qwen3_vl} & API & n/r \\
Qwen3-VL-Flash~\cite{qwen3_vl} & API & n/r \\
Qwen3-VL-8B-Instruct~\cite{qwen3_vl} & weights & 8B \\
Qwen3-VL-32B-Instruct~\cite{qwen3_vl} & weights & 32B \\
Pixtral-12B~\cite{pixtral} & weights & 12B \\
GPT-4.1-mini~\cite{openai_gpt41mini} & API & n/r \\
GPT-5.4-mini~\cite{openai_gpt54mini} & API & n/r \\
Claude Haiku 4.5~\cite{anthropic_claude} & API & n/r \\
\bottomrule
\end{tabular}

\end{table}


\FloatBarrier
\endgroup
\section{Case Study by Error Type}
\label{appendix:error-analysis}

These six cases compare model predictions and judge decisions with museum records. They show where attributions follow an object's style or a familiar name rather than the record of its making, and where judge scoring is inconsistent.
\par\smallskip

\clearpage
\noindent
\begin{minipage}{\columnwidth}
\subsection{Case Study A: Cross-Cultural Misattribution}
\label{sec:case_A}

\textbf{Object ID: 1055\_Butter Pat}

\begin{figure}[H]
    \centering
    \includegraphics[width=0.32\linewidth,height=0.70\textheight,keepaspectratio]{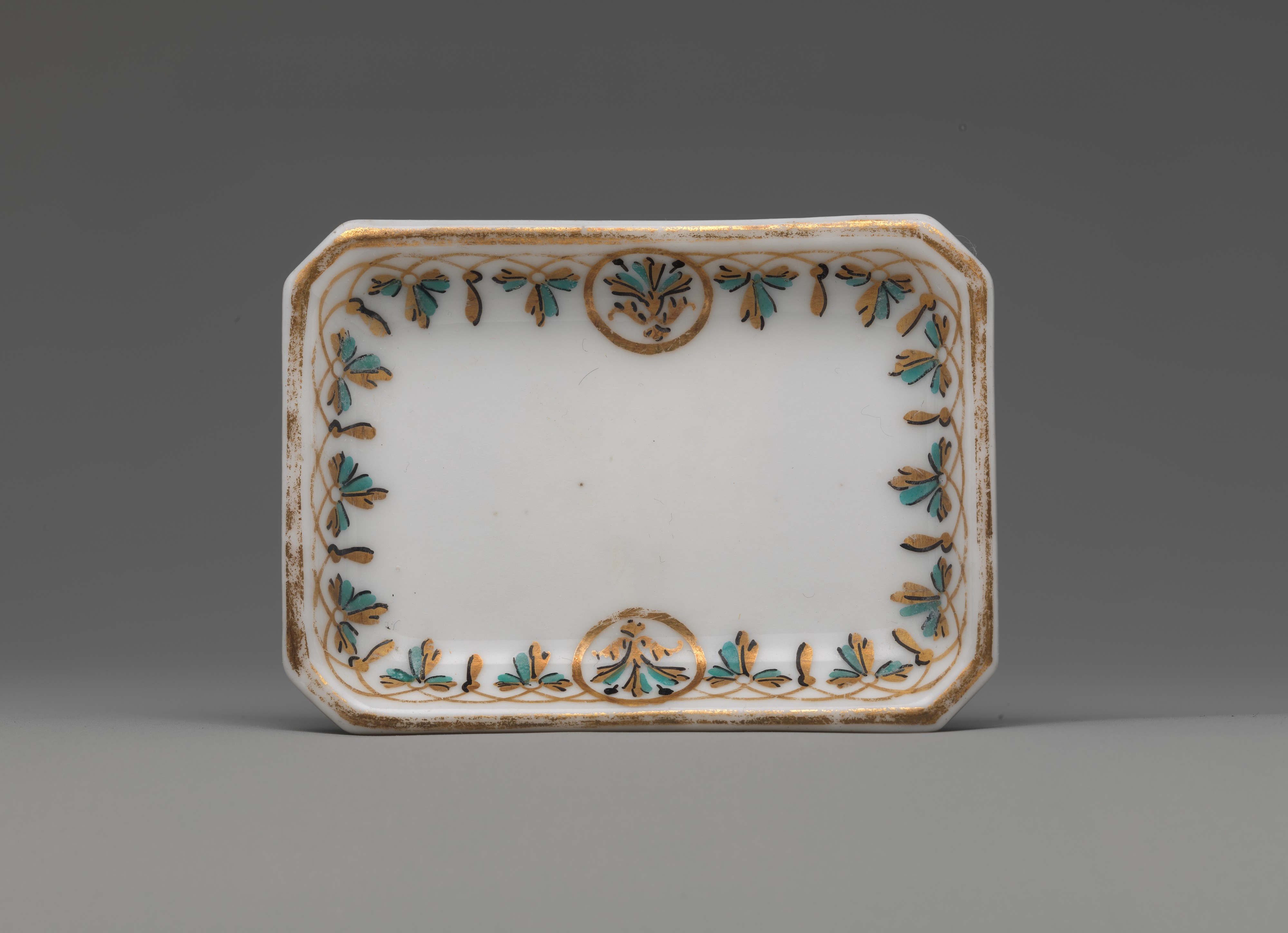}
    \caption{Object 1055\_Butter Pat.}
\end{figure}
\end{minipage}
\par

\begin{samepage}
\textbf{Ground Truth:}
\begin{itemize}
    \item Title: Butter Pat
    \item Culture: American
    \item Period: 1885; 19th century
    \item Origin: United States; America
    \item Creator: Union Porcelain Works
\end{itemize}
\end{samepage}

\textbf{Representative Predictions:}
\begin{itemize}
    \item Claude Haiku 4.5: French or European, late 18th century
    \item GPT-4.1-mini: European, France, 18th century
    \item Qwen-VL-Max: Japanese, Meiji Period
    \item Pixtral-12B: Chinese, Qing Dynasty
\end{itemize}

\textbf{Analysis:}

Claude Haiku 4.5 and GPT-4.1-mini assign European or French culture and eighteenth-century dates, conflicting with the recorded American culture and 1885 date. Qwen-VL-Max and Pixtral-12B instead assign Japanese and Chinese cultures, respectively, which also differ from the museum record.

\FloatBarrier
\clearpage
\noindent
\begin{minipage}{\columnwidth}
\subsection{Case Study B: Culture and Maker Misattribution}
\label{sec:case_B}

\textbf{Object ID: 1513\_Celery vase}

\begin{figure}[H]
    \centering
    \includegraphics[width=0.22\linewidth,height=0.70\textheight,keepaspectratio]{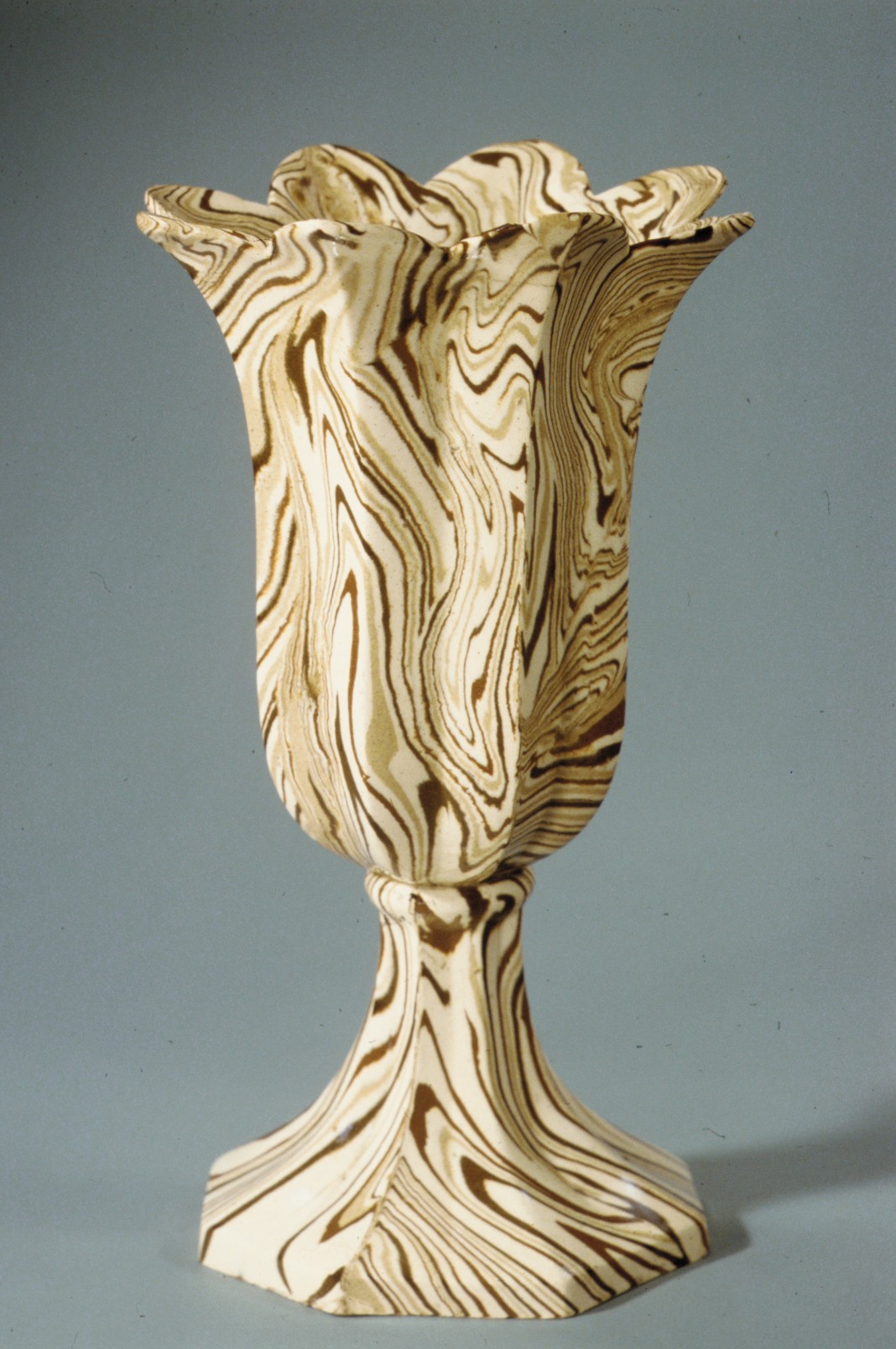}
    \caption{Object 1513\_Celery vase.}
\end{figure}
\end{minipage}
\par

\begin{samepage}
\textbf{Ground Truth:}
\begin{itemize}
    \item Title: Celery vase
    \item Culture: American
    \item Period: 1849--1858; mid-19th century
    \item Origin: United States; America
    \item Creator: United States Pottery Company
\end{itemize}
\end{samepage}

\textbf{Representative Predictions:}
\begin{itemize}
    \item GPT-4.1-mini: Dutch, Delftware workshop
    \item Qwen-VL-Max: English, Wedgwood
    \item Qwen3-VL-Plus: British, Staffordshire
    \item Claude Haiku 4.5: European modernist
\end{itemize}

\textbf{Analysis:}

The four displayed predictions assign European cultural contexts to an object recorded as American. GPT-4.1-mini's Delftware workshop and Qwen-VL-Max's Wedgwood attribution differ from the recorded United States Pottery Company. Qwen3-VL-Plus assigns British culture and Staffordshire, while Claude Haiku 4.5 describes a European modernist context.

\FloatBarrier
\clearpage
\noindent
\begin{minipage}{\columnwidth}
\subsection{Case Study C: Object Recognition without Cultural Attribution}
\label{sec:case_C}

\textbf{Object ID: 42\_Andiron}

\begin{figure}[H]
    \centering
    \includegraphics[width=0.25\linewidth,height=0.70\textheight,keepaspectratio]{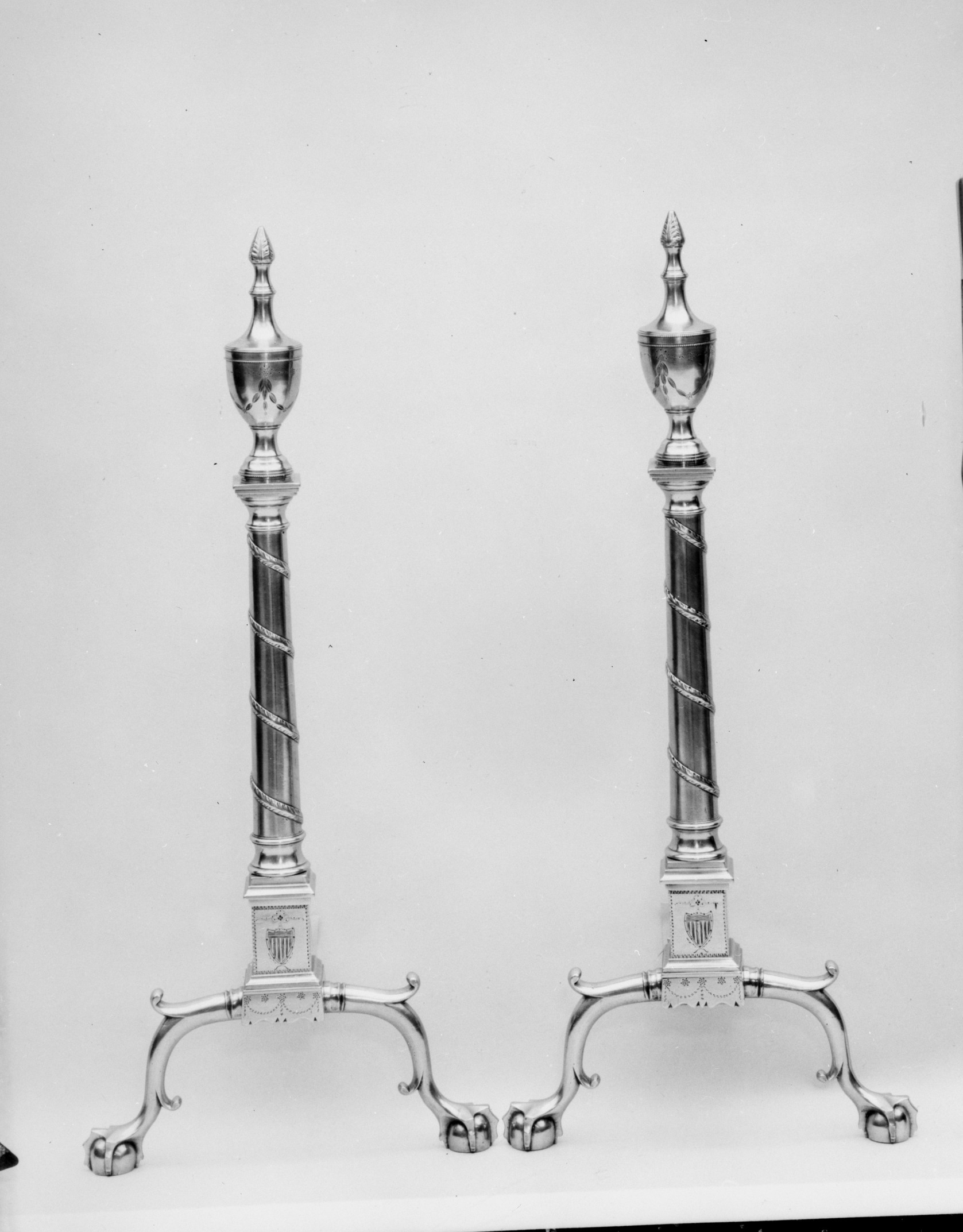}
    \caption{Object 42\_Andiron.}
\end{figure}
\end{minipage}
\par

\begin{samepage}
\textbf{Ground Truth:}
\begin{itemize}
    \item Title: Andiron
    \item Culture: American
    \item Period: 1795--1810; late 18th century; early 19th century
    \item Origin: Not specified (empty reference array)
    \item Creator: Not specified (empty reference array)
\end{itemize}
\end{samepage}

\textbf{Representative Predictions:}
\begin{itemize}
    \item Pixtral-12B: Pair of Empire Style Andirons; Western; Early 19th Century; France; Unknown French Workshop (DS-J: title and period compatible; culture, origin, and creator incompatible)
    \item Claude Haiku 4.5: Pair of Andirons with Fireplace Tools; American; Late 19th--Early 20th century; United States (DS-J: period incompatible)
    \item GPT-4.1-mini: Pair of Neoclassical Andirons; American; Late 18th to early 19th century; United States (DS-J: all five fields compatible)
\end{itemize}

\textbf{Analysis:}

Pixtral-12B identifies andirons and gives a compatible early-nineteenth-century date, but its Western culture label differs from the recorded American culture. Its France origin and French workshop attribution add specific information where the references are empty. DS-J marks its title and period compatible and the other three fields incompatible. In the displayed alternatives, Claude Haiku 4.5 has an incompatible period. DS-J marks all five GPT-4.1-mini fields compatible, although its origin is the definite answer United States against an empty reference; the accompanying reason identifies that mismatch. The stored score and reason therefore disagree.

\FloatBarrier
\clearpage
\noindent
\begin{minipage}{\columnwidth}
\subsection{Case Study D: Broad Categorisation without Specific Identification}
\label{sec:case_D}

\textbf{Object ID: 0f097d4a-4ca1-40fd-b562-ab41a411aff1}

\begin{figure}[H]
    \centering
    \includegraphics[width=0.17\linewidth,height=0.70\textheight,keepaspectratio]{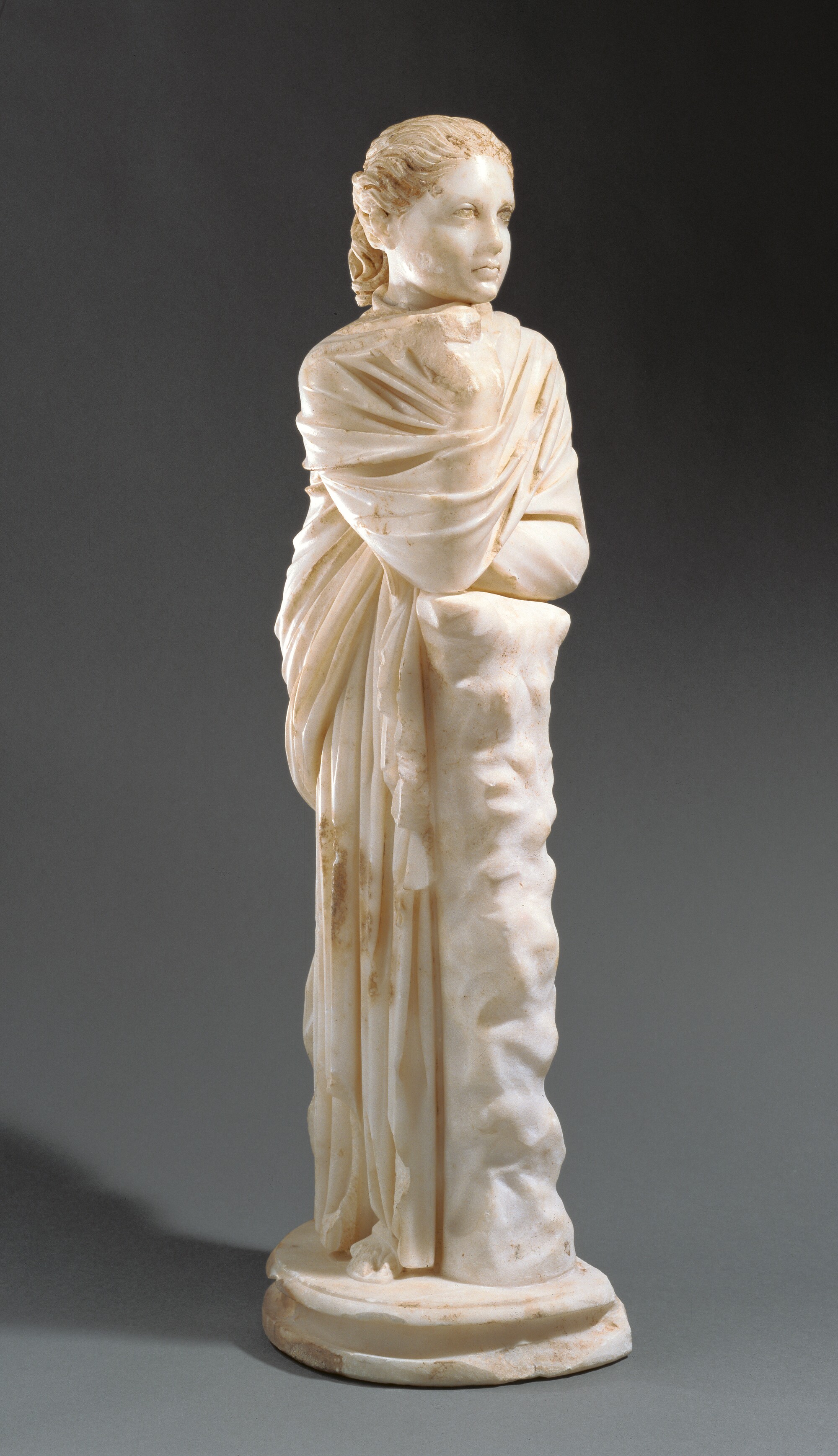}
    \caption{Object 0f097d4a-4ca1-40fd-b562-ab41a411aff1.}
\end{figure}
\end{minipage}
\par

\begin{samepage}
\textbf{Ground Truth:}
\begin{itemize}
    \item Title: Statue of a Muse
    \item Culture: Roman
    \item Period: Not specified (empty reference array)
    \item Origin: Not specified (empty reference array)
    \item Creator: Unknown
\end{itemize}
\end{samepage}

\textbf{Additional Context (Museum Description):}
\begin{itemize}
    \item Identified as a Muse (likely Polyhymnia)
    \item Associated with Roman imperial architectural decoration
    \item Originally part of a sculptural group
\end{itemize}

\textbf{Representative Predictions:}
\begin{itemize}
    \item GPT-5.4-mini: Funerary statue of a young woman; Roman; Imperial Roman, 1st century CE; Eastern Mediterranean, likely Syria or Asia Minor
    \item Pixtral-12B: Standing Female Figure; Roman; 1st--2nd century AD; Italy or North Africa
    \item Qwen3-VL-Plus: Statue of a draped female figure (possibly a Muse or nymph); Roman; Early Imperial period, 1st century CE; Rome or central Italy
\end{itemize}

\textbf{Analysis:}

The three displayed predictions report Roman culture but differ in title. GPT-5.4-mini calls the object a funerary statue of a young woman, and Pixtral-12B gives the broad title \emph{Standing Female Figure}; neither reproduces the recorded identification as a Muse. Qwen3-VL-Plus includes \emph{Muse} as one tentative alternative. The predictions also supply periods and origins despite empty reference arrays for those fields. The museum description provides additional context but does not fill empty references under the annotation rules.

\FloatBarrier
\clearpage
\noindent
\begin{minipage}{\columnwidth}
\subsection{Case Study E: Over-Specific Metadata Predictions}
\label{sec:case_E}

\textbf{Object ID: 333\_Basin}

\begin{figure}[H]
    \centering
    \includegraphics[width=0.25\linewidth,height=0.70\textheight,keepaspectratio]{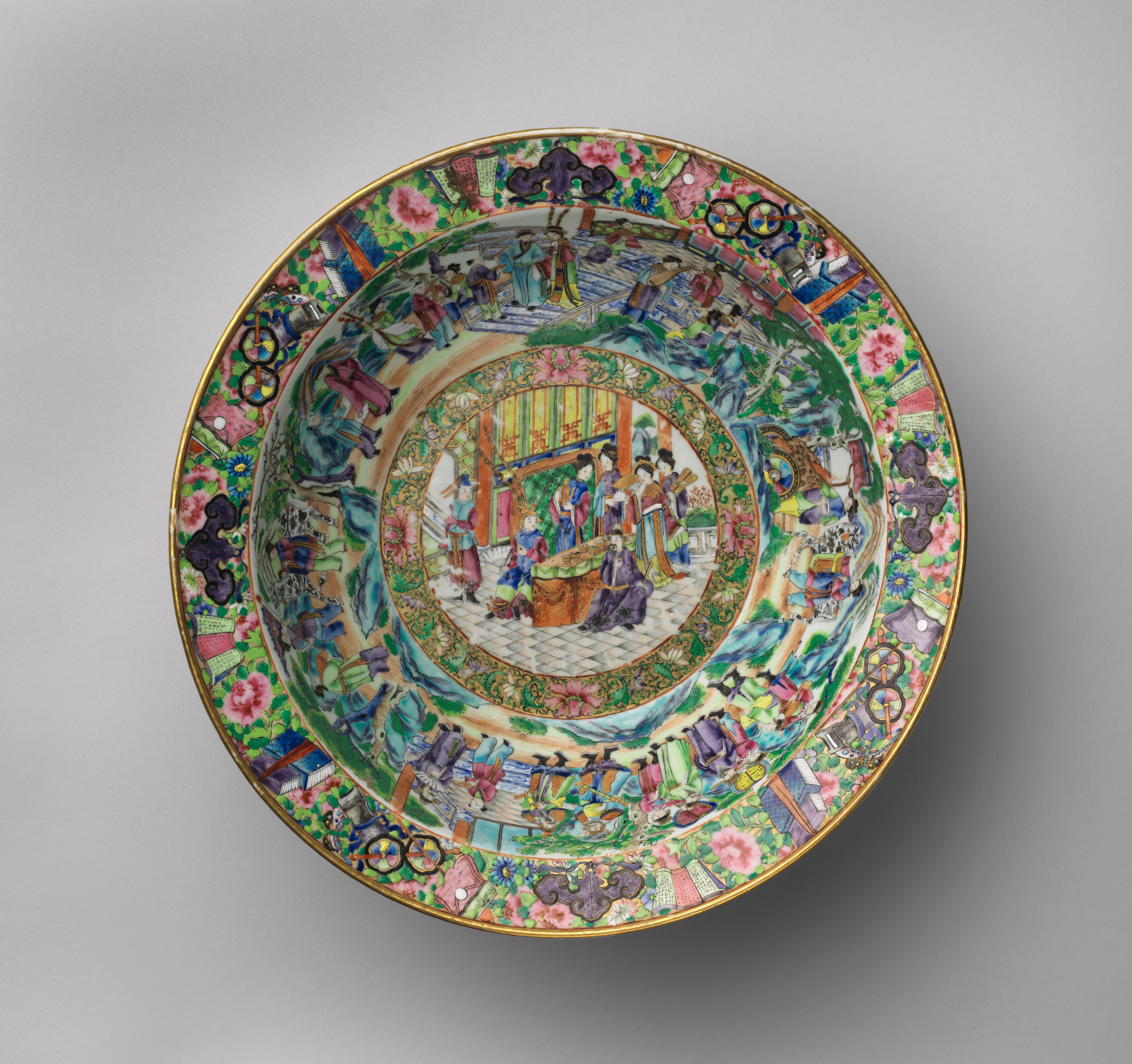}
    \caption{Object 333\_Basin.}
\end{figure}
\end{minipage}
\par

\begin{samepage}
\textbf{Ground Truth:}
\begin{itemize}
    \item Title: Basin
    \item Culture: Chinese
    \item Period: 1825--1845; 19th century
    \item Origin: Not specified (empty reference array)
    \item Creator: Not specified (empty reference array)
\end{itemize}
\end{samepage}

\textbf{Representative Predictions:}
\begin{itemize}
    \item Qwen3-VL-Plus: Cantonese Export Porcelain Bowl with `The Eight Immortals' and Courtly Scene; Chinese (Cantonese); Qing Dynasty, Guangxu reign (1875--1908); Guangzhou (Canton), China; Workshop of the Thirteen Factories
    \item Claude Haiku 4.5: Dish with Narrative Scenes and Floral Borders (Famille Verte); Chinese; Qing Dynasty, Kangxi period (1662--1722); Jingdezhen, Jiangxi Province, China; Jingdezhen Imperial or commercial kiln workshop
    \item GPT-5.4-mini: Famille en visite de printemps (Spring outing) bowl; Chinese; Qing dynasty, late 18th to 19th century; Jingdezhen, Jiangxi Province, China; specific workshop not identified
\end{itemize}

\textbf{Analysis:}

The three displayed predictions report Chinese culture, matching the reference. Qwen3-VL-Plus assigns the Guangxu reign (1875--1908), and Claude Haiku 4.5 the Kangxi period (1662--1722); both differ from the recorded 1825--1845 date. Qwen3-VL-Plus also names the Eight Immortals, a detail not documented in the supplied metadata. The outputs add production locations or workshop attributions where the corresponding references are empty. GPT-5.4-mini gives a broad compatible period and leaves the specific workshop unidentified. DS-J also marks its definite origin, Jingdezhen, compatible by citing uncertainty expressed in the creator field, although the rubric requires independent field decisions.

\FloatBarrier
\clearpage
\noindent
\begin{minipage}{\columnwidth}
\subsection{Case Study F: Title and Creator Mismatch}
\label{sec:case_F}

\textbf{Object ID: 2b6e224c-686a-4b43-aa5a-1ef5520ef0ef}

\begin{figure}[H]
    \centering
    \includegraphics[width=0.28\linewidth,height=0.70\textheight,keepaspectratio]{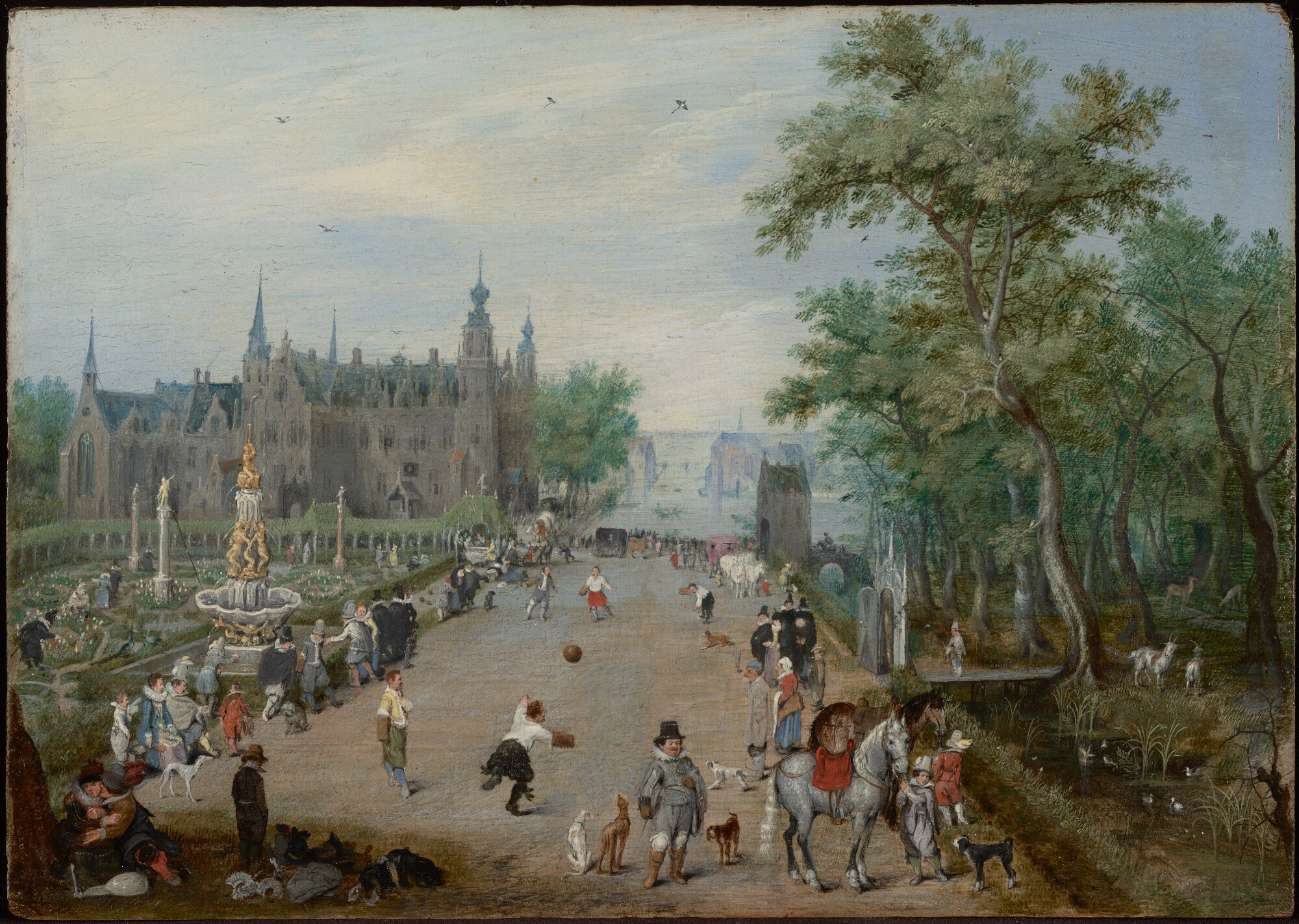}
    
    \vspace{0.2em}
    
    \begin{minipage}{0.25\linewidth}
        \centering
        \includegraphics[width=\linewidth,height=0.70\textheight,keepaspectratio]{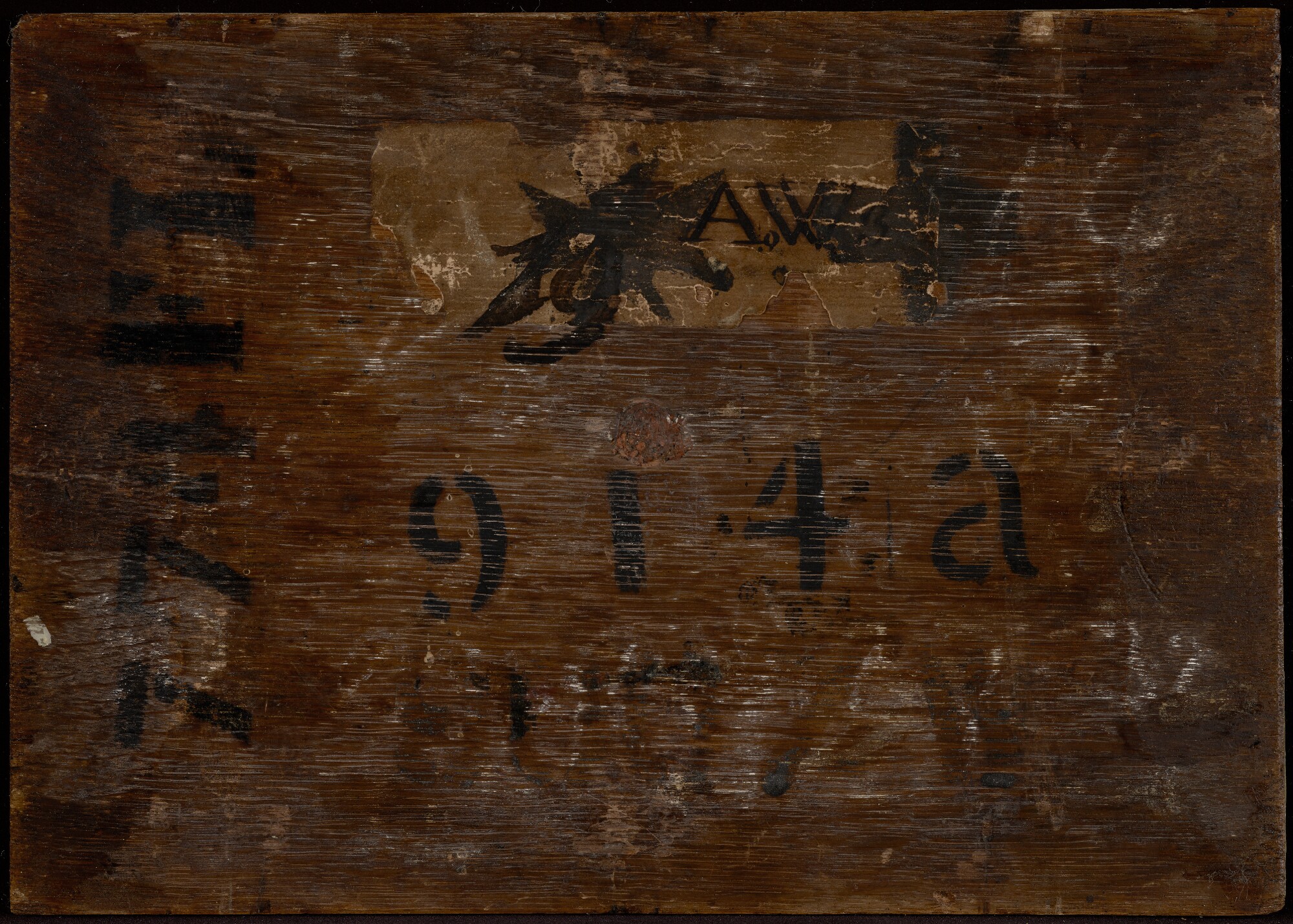}
    \end{minipage}
    \hfill
    \begin{minipage}{0.25\linewidth}
        \centering
        \includegraphics[width=\linewidth,height=0.70\textheight,keepaspectratio]{figures/caseF-2.jpg}
    \end{minipage}
    
    \caption{Object 2b6e224c-686a-4b43-aa5a-1ef5520ef0ef. Top: painting; bottom: back (left) and framed view (right).}
\end{figure}
\end{minipage}
\par

\begin{samepage}
\textbf{Ground Truth:}
\begin{itemize}
    \item Title: A Ball Game Before a Country Palace
    \item Culture: Dutch
    \item Period: 17th century; Early 17th century; Baroque
    \item Origin: Netherlands; Dutch Republic
    \item Creator: Adriaen van de Venne; Adriaen van de Venne (Dutch, 1589--1662)
\end{itemize}
\end{samepage}

\textbf{Additional Context (Museum Description):}
\begin{itemize}
    \item Outdoor social scene featuring a ball game with multiple figures in a landscaped setting.
    \item Includes a formal garden, a palace-like structure, and surrounding natural elements.
    \item Contains narrative details such as leisure activities, animals, and symbolic objects.
    \item Likely part of a seasonal landscape series, associated with summer.
    \item Characterised by small-scale figures and detailed anecdotal composition.
\end{itemize}

\textbf{Illustrative Prediction (Pixtral-12B):}
\begin{itemize}
    \item Title: View of the Courtyard of the Amsterdam City Hall
    \item Culture: Dutch
    \item Period: Golden Age
    \item Origin: Amsterdam
    \item Creator: Not sure, possibly a follower of Pieter Saenredam or a similar artist
\end{itemize}

\textbf{Analysis:}

Pixtral-12B's Dutch culture matches the reference, but its title names a different scene and its suggested maker differs from Adriaen van de Venne. Full-run DS-J accepts the culture and \emph{Golden Age} period and rejects the title, origin, and creator. Its origin reason rejects Amsterdam because it is a city rather than the recorded country or region, although the rubric permits a compatible city to match a broader origin.

\end{document}

%% file: preamble.tex
\usepackage[T1]{fontenc}
\usepackage[utf8]{inputenc}
\usepackage{microtype}
\usepackage{adjustbox,array,placeins,listings,pifont,float,makecell,siunitx,colortbl,multirow}
\definecolor{PothenForest}{HTML}{21482D}
\definecolor{PothenTerracotta}{HTML}{BC603D}
\definecolor{PothenWarmGray}{HTML}{948E80}
\newcommand{\cmark}{\textcolor{PothenForest}{\ding{51}}}
\newcommand{\xmark}{\textcolor{PothenTerracotta}{\ding{55}}}
\newcommand{\pmark}{\textcolor{PothenWarmGray}{\ding{108}}}